%% file: Full_nestedkmeans.tex
\documentclass{article}
\usepackage[utf8]{inputenc} 
\usepackage{amstext, amsmath,latexsym,amsbsy,amssymb}
\usepackage{hyperref}
\usepackage{amsfonts}
\usepackage[accepted]{icml2017} 
\usepackage{array}
\usepackage{esint}
\usepackage[pdftex]{graphicx}
\usepackage{amsfonts,amsmath,amssymb,amsthm}
\usepackage{epsf}
\usepackage{graphics}
\usepackage{psfrag}
\usepackage{times}
\usepackage{epsfig}
\usepackage{subfig}
\usepackage[titletoc]{appendix}
\usepackage{multirow}
\usepackage{hhline}
\usepackage{color}
\usepackage{bbm}
\usepackage{algorithm}
\usepackage{algorithmic}
\usepackage{booktabs}
\usepackage{hyperref}

\usepackage{colortbl} 
\definecolor{header_color}{rgb}{0.74,0.88,0.91}
\definecolor{even_color}{rgb}{0.9,0.9,0.9}
\definecolor{subheader_color}{rgb}{0.85,0.93,0.95}
\definecolor{childheader_color}{rgb}{1.0,0.93,0.87}

\definecolor{ccolor_best}{rgb}{1.0,0.9,0.9}
\definecolor{ccolor_wrong}{rgb}{1.0,0.85,0.85}

\makeatother

\RequirePackage{natbib}
\long\def\comment#1{}
\renewcommand\vec[1]{\ensuremath\boldsymbol{#1}}

\newcommand{\Hcal}{\ensuremath{\mathcal{H}}}
\newcommand{\Hbold}{\ensuremath{\boldsymbol{H}}}

\comment{
\setlength{\topmargin}{0 in}
\setlength{\textwidth}{5.5 in}
\setlength{\textheight}{8 in}
\setlength{\oddsidemargin}{0.5 in}
}

\theoremstyle{plain}

\newtheorem{innercustomthm}{Theorem}
\newenvironment{customthm}[1]
  {\renewcommand\theinnercustomthm{#1}\innercustomthm}
  {\endinnercustomthm}
  
  \newtheorem{innercustomlem}{Lemma}
\newenvironment{customlem}[1]
  {\renewcommand\theinnercustomlem{#1}\innercustomlem}
  {\endinnercustomlem}

  \newtheorem{innercustomprop}{Proposition}
\newenvironment{customprop}[1]
  {\renewcommand\theinnercustomprop{#1}\innercustomprop}
  {\endinnercustomprop}
\theoremstyle{plain}

\newtheorem{theorem}{Theorem}

\numberwithin{theorem}{section}

\numberwithin{proposition}{section}

\numberwithin{lemma}{section}

\numberwithin{definition}{section}

\numberwithin{condition}{section}

\numberwithin{problem}{section}

\numberwithin{corollary}{section}

\numberwithin{assumption}{section}

\numberwithin{example}{section}

\numberwithin{conjecture}{section}

\theoremstyle{definition}

\numberwithin{remark}{section}

\begin{document}
\twocolumn[
\icmltitle{Multilevel Clustering via Wasserstein Means}


\icmlsetsymbol{equal}{*}

\begin{icmlauthorlist}
\icmlauthor{Nhat Ho}{mi}
\icmlauthor{XuanLong Nguyen}{mi}
\icmlauthor{Mikhail Yurochkin}{mi}
\icmlauthor{Hung Hai Bui}{ado}
\icmlauthor{Viet Huynh}{dea}
\icmlauthor{Dinh Phung}{dea}
\end{icmlauthorlist}

\icmlaffiliation{mi}{Department of Statistics, University of Michigan, USA.}
\icmlaffiliation{dea}{Center for Pattern Recognition and Data Analytics (PRaDA), Deakin University, Australia}
\icmlaffiliation{ado}{Adobe Research.}

\icmlcorrespondingauthor{Nhat Ho}{minhnhat@umich.edu}

\icmlkeywords{Multilevel Clustering, Optimal Transport}

\vskip 0.3in
]



\printAffiliationsAndNotice{}

\begin{abstract} 
We propose a novel approach to the problem of multilevel clustering, which aims to 
simultaneously partition data in each group and discover grouping patterns among groups
in a potentially large hierarchically structured corpus of data.
Our method involves a joint optimization formulation over several spaces of discrete probability measures,
which are endowed with Wasserstein distance metrics. We propose a number of variants
of this problem, which admit fast optimization algorithms, by exploiting the connection to the
problem of finding Wasserstein barycenters.  Consistency properties are established
for the estimates of both local and global clusters.
Finally, experiment results with both synthetic and real data are presented to demonstrate the 
flexibility and scalability of the proposed approach. \footnote{Code is available at \url{https://github.com/moonfolk/Multilevel-Wasserstein-Means}}
\end{abstract} 
\section{Introduction} \label{Section:introduction}
\input{arxiv_nested_introduction}

\section{Background} \label{Section:prelim}
\input{arxiv_nested_prelim}

\section{Clustering with multilevel structure data} \label{Section:multilevel_Wasserstein}
\input{arxiv_nested_multilevel}

\section{Consistency results} \label{Section:consistency_multilevel_Kmeans}
\input{arxiv_nested_consistency}

\section{Empirical studies} \label{Section:data_analysis}
\input{arxiv_nested_simulation}
\vspace{-6pt}
\section{Discussion} \label{Section:discussion}
We have proposed an optimization based approach to multilevel clustering 
using Wasserstein metrics. There are several possible directions for extensions.
Firstly, we have only considered continuous data; it is of interest to extend 
our formulation to discrete data. Secondly, our 
method requires knowledge of the numbers of clusters both in local and global clustering. 
When these numbers are unknown, it seems reasonable to incorporate penalty
on the model complexity. Thirdly,  our formulation does not directly account for the ``noise'' distribution away from the 
(Wasserstein) means.  To improve the robustness, it may be desirable to
make use of the first-order Wasserstein metric instead of the second-order one. 
Finally, we are interested in extending our approach to richer settings of 
hierarchical data, such as one when group level-context is available. 
\vspace{-6pt}
\paragraph{Acknowledgement.} This research is supported in part by grants
NSF CAREER DMS-1351362, NSF CNS-1409303, the Margaret and Herman Sokol 
Faculty Award and research gift from Adobe Research (XN). 
DP gratefully acknowledges the partial support from the 
Australian Research Council (ARC) and AOARD (FA2386-16-1-4138).

\newpage
\bibliography{Nhat,NPB,Nguyen,MY_ref}

\newpage
\begin{center}
\textbf{\Large Appendix A}
\end{center}
\input{arxiv_nested_appendix_a}

\begin{center}
\textbf{\Large Appendix B}
\end{center}
\input{arxiv_nested_appendix_b}

\begin{center}
\textbf{\Large Appendix C}
\end{center}
\input{arxiv_nested_appendix_c}

\begin{center}
\textbf{\Large Appendix D}
\end{center}
\input{arxiv_nested_appendix_d}
\end{document}

%% file: arxiv_nested_introduction.tex
In numerous applications in engineering and sciences, data are often organized in
a multilevel structure. For instance, a typical structural view of text data in machine learning
is to have words grouped into documents, documents are grouped into 
corpora. A prominent strand of modeling and algorithmic works in the past couple decades
has been to discover latent multilevel structures from these hierarchically structured data. 
For specific clustering tasks, one may be interested in simultaneously partitioning the data in each 
group (to obtain local clusters) and partitioning a collection of data groups (to obtain global clusters). 
Another concrete example is the problem of clustering images (i.e., global clusters) where each image 
contains partions of multiple annotated regions (i.e., local clusters) \citep{Oliva-2001}. 
While hierachical clustering techniques may be employed to find a tree-structed clustering 
given a collection of data points, they are not applicable to discovering the nested structure of multilevel data.
Bayesian hierarchical models provide a powerful approach, exemplified by influential works such as
\cite{Blei-etal-03,Pritchard-etal-00,Teh-etal-06}. More specific to the simultaneous and multilevel
clustering problem, we mention the paper of \cite{Rodriguez-etal-08}. In this interesting work,
a Bayesian nonparametric model, namely the nested Dirichlet process (NDP) model,
was introduced that enables
the inference of clustering of a collection of probability distributions from which different 
groups of data are drawn. With suitable extensions, this modeling framework
has been further developed for simultaneous multilevel clustering, see for instance, 
\citep{Wulsin-2016,Vu-2014,Viet-2016}. 

The focus of this paper is on the multilevel clustering problem motivated in the aforementioned
modeling works, but we shall take a purely optimization approach. 
We aim to formulate optimization problems that enable the discovery of multilevel clustering 
structures hidden in grouped data. Our technical approach is
inspired by the role of optimal transport distances in hierarchical modeling and clustering problems.
The optimal transport distances, also known as Wasserstein distances \citep{Villani-03},
have been shown to be the natural distance metric for the convergence theory of latent
mixing measures arising in both mixture models \citep{Nguyen-13} and hierarchical models \citep{Nguyen-2016}.
They are also intimately connected to the problem of clustering --- this relationship goes
back at least to the work of \citep{Pollard-1982}, where it is pointed out that the well-known 
K-means clustering algorithm can be directly linked to the quantization problem --- the problem
of determining an optimal finite discrete probability measure that minimizes its 
second-order Wasserstein distance from the empirical distribution of given data \citep{Graf-2000}.

If one is to perform simultaneous K-means clustering for hierarchically grouped data, both at the global
level (among groups), and local level (within each group), then this can be achieved by a joint optimization
problem defined with suitable notions of Wasserstein distances inserted into the objective
function. In particular, multilevel clustering requires the optimization in the space of
probability measures defined in \emph{different} levels of abstraction, including the space of
measures of measures on the space of grouped data.
Our goal, therefore, is to formulate this optimization precisely, to develop algorithms
for solving the optimization problem efficiently, and to make sense of the obtained solutions
in terms of statistical consistency. 

The algorithms that we propose address directly a multilevel clustering problem 
formulated from a purely optimization viewpoint, but they may also be taken as a fast approximation to
the inference of latent mixing measures that arise in the nested Dirichlet process of \citep{Rodriguez-etal-08}.
From a statistical viewpoint, we shall establish a consistency theory for our multilevel clustering
problem in the manner achieved for K-means clustering \citep{Pollard-1982}. From a computational viewpoint,
quite interestingly, we will be able to explicate and exploit the connection betwen our optimization
and that of finding the Wasserstein barycenter \citep{Carlier-2011}, an interesting
computational problem that have also attracted much recent interests, e.g., \citep{Cuturi-2014}.

In summary, the main contributions offered in this work include (i) 
a new optimization formulation to the multilevel clustering problem using Wasserstein
distances defined on different levels of the hierarchical data structure; (ii) 
fast algorithms by exploiting the connection of our formulation to the Wasserstein 
barycenter problem; (iii) consistency theorems established for proposed
estimates under very mild condition of data's distributions; 
(iv) several flexibile alternatives by introducing constraints that encourage the borrowing
of strength among local and global clusters, and (v) finally, demonstration of 
efficiency and flexibility of our approach in a number of simulated and real data sets.  

The paper is organized as follows. Section \ref{Section:prelim} provides preliminary 
background on Wasserstein distance, Wasserstein barycenter, and the connection between 
K-means clustering and the quantization problem. Section 
\ref{Section:multilevel_Wasserstein} presents
several optimization formulations of the multilevel clusering problem, and the algorithms
for solving them. Section
\ref{Section:consistency_multilevel_Kmeans} establishes consistency results of the estimators
introduced in Section \ref{Section:consistency_multilevel_Kmeans}. Section 
\ref{Section:data_analysis} presents careful simulation studies with both synthetic and real 
data. Finally, we conclude the paper with a discussion in Section \ref{Section:discussion}.
Additional technical details, including all proofs, are given in the Supplement.

%% file: arxiv_nested_prelim.tex

For any given subset $\Theta \subset \mathbb{R}^{d}$, let $
\mathcal{P}(\Theta)$ denote the space of Borel probability measures on $\Theta$. 
The Wasserstein space of order $r \in [1,\infty)$ of probability measures on $\Theta$ is defined as
$\mathcal{P}_{r}(\Theta)=\biggr\{G \in \mathcal{P}(\Theta): \ \int \limits {\|x\|^{r}}dG(x)<\infty \biggr\}$,
where $\|.\|$ denotes Euclidean metric in $\mathbb{R}^{d}$. Additionally, for any $k \geq 
1$ the probability simplex is denoted by $\Delta_{k}=\left\{u \in \mathbb{R}^{k}: \ u_{i} \geq 0, \ 
\sum \limits_{i=1}^{k}{u_{i}}=1 \right\}$. Finally, let $\mathcal{O}_{k}(\Theta)$ 
(resp., $\mathcal{E}_{k}(\Theta)$) 
be the set of probability 
measures with at most (resp., exactly) $k$ support points in $\Theta$.

\paragraph{Wasserstein distances}
For any elements $G$ and $G'$ in $\mathcal{P}_{r}(\Theta)$ where $r \geq 1$, the 
Wasserstein distance of order $r$ between $G$ and $G'$ is defined as (cf. \citep{Villani-03}):
\vspace{-6pt}
\begin{eqnarray}
W_{r}(G,G')=\biggr(\mathop {\inf }\limits_{\pi \in \Pi(G,G')}{\int \limits_{\Theta^{2}}{\|x-y\|^{r}}d\pi(x,y)}\biggr)^{1/r} \nonumber
\end{eqnarray}
where $\Pi(G,G')$ is the set of all probability measures on $\Theta \times \Theta$ that have 
marginals $G$ and $G'$. 
In words, $W_r^r(G,G')$ is
the optimal cost of moving mass from $G$ to $G'$, where the cost of moving unit mass is
proportional to $r$-power of Euclidean distance in $\Theta$. 
When $G$ and $G'$ are two discrete measures with finite number 
of atoms, fast computation of $W_r(G,G')$ can be achieved (see, e.g., \cite{Cuturi-2013}). 
The details of this are deferred to the Supplement.
 
By a recursion of concepts, we can speak of measures of measures, and define a suitable distance metric
on this abstract space: 
the space of Borel measures on $\mathcal{P}_{r}(\Theta)$, to be denoted by
$\mathcal{P}_{r}(\mathcal{P}_{r}(\Theta))$. This is also a Polish space (that is,
complete and separable metric space) as $
\mathcal{P}_{r}(\Theta)$ is a Polish space. It will be endowed with a Wasserstein metric of 
order $r$ that is induced by a metric $W_{r}$ on $\mathcal{P}_{r}(\Theta)$ as follows (cf. 
Section 3 of \cite{Nguyen-2016}): for any $\mathcal{D},\mathcal{D'} \in \mathcal{P}_r(\mathcal{P}_r(\Theta))$
\vspace{-6pt}
\begin{eqnarray}
W_{r}(\mathcal{D},\mathcal{D}'):=\biggr(\mathop {\inf }{\int \limits_{\mathcal{P}_{r}(\Theta)^{2}}{W_{r}^{r}(G,G')}d\pi(G,G')}\biggr)^{1/r} \nonumber
\end{eqnarray}
where the infimum in the above ranges over all $\pi \in \Pi(\mathcal{D},\mathcal{D}')$ 
such that $\Pi(\mathcal{D},\mathcal{D}')$ is the set of all probability measures on $
\mathcal{P}_{r}(\Theta) \times \mathcal{P}_{r}(\Theta)$ that has marginals $\mathcal{D}
$ and $\mathcal{D}'$. In words, $W_r(\mathcal{D},\mathcal{D'})$ corresponds to 
the optimal cost of moving mass from $\mathcal{D}$ to $\mathcal{D'}$, where
the cost of moving unit mass in its space of support $\mathcal{P}_r(\Theta)$ 
is proportional to the $r$-power of the $W_r$ distance in $\mathcal{P}_r(\Theta)$.
Note a slight notational abuse --- $W_r$ is used for both
$\mathcal{P}_r(\Theta)$ and $\mathcal{P}_r(\mathcal{P}_r(\Theta))$, but
it should be clear which one is being used from context.

\paragraph{Wasserstein barycenter}
Next, we present a brief overview of Wasserstein barycenter problem, first
studied by \citep{Carlier-2011} and subsequentially many others (e.g., \citep{Benamou-15, Solomon-15, Alvarez-16}). 
Given probability measures 
$P_{1}, P_{2}, \ldots, P_{N} \in \mathcal{P}_{2}(\Theta)$ for $N \geq 1$, their 
Wasserstein barycenter $\overline{P}_{N,\lambda}$ is such that
\vspace{-6pt}
\begin{eqnarray}
\overline{P}_{N,\lambda}=\mathop {\arg \min}\limits_{P \in \mathcal{P}_{2}(\Theta)}{\sum \limits_{i=1}^{N}{\lambda_{i}W_{2}^{2}(P,P_{i})}} \label{eqn:Wasserstein_barycenter}
\end{eqnarray} 
where $\lambda \in \Delta_{N}$ denote weights associated with $P_{1},\ldots,P_{N}$.
When $P_{1},\ldots, P_{N}$ are discrete measures with finite number of atoms and the 
weights $\lambda$ are uniform, it was shown by \citep{Anderes-2015}
that the problem of finding Wasserstein barycenter $\overline{P}
_{N,\lambda}$ over the space $\mathcal{P}_{2}(\Theta)$ in 
\eqref{eqn:Wasserstein_barycenter} is reduced to search only over a much simpler space 
$\mathcal{O}_{l}(\Theta)$ 
where $l=\sum \limits_{i=1}
^{N}{s_{i}-N+1}$ and $s_{i}$ is the number of components of $P_{i}$ for all $1 \leq i \leq 
N$. 
Efficient algorithms for finding local solutions of the Wasserstein barycenter problem 
over $\mathcal{O}_{k}(\Theta)$ for some $k \geq 1$ have been studied recently in 
\citep{Cuturi-2014}. These algorithms will prove to be a useful building block for 
our method as we shall describe in the sequel. 
The notion of Wasserstein barycenter has been utilized for approximate Bayesian inference
\citep{Sanvesh-aistats}.

\paragraph{K-means as quantization problem}
The well-known $K$-means clustering algorithm can be viewed as solving
an optimization problem that arises in the problem of quantization, a simple but very useful connection 
\citep{Pollard-1982, Graf-2000}. The connection is the following.
Given $n$ unlabelled samples $Y_{1},\ldots,Y_{n} \in \Theta$. Assume that these data are associated
with at most $k$ clusters where $k \geq 1$ is some given number. The $K$-means problem finds the set $S$ 
containing at most $k$ elements $\theta_{1},\ldots, \theta_{k} \in \Theta$ that minimizes 
the following objective
\vspace{-6pt}
\begin{eqnarray}
\mathop {\inf }\limits_{S : |S| \leq k}{\dfrac{1}{n}\sum \limits_{i=1}^{n}{d^{2}(Y_{i},S)}}. \label{eqn:original_Kmeans}
\end{eqnarray}
Let $P_{n}=\dfrac{1}{n}\sum \limits_{i=1}^{n}{\delta_{Y_{i}}}$ be the empirical measure of data 
$Y_{1},\ldots,Y_{n}$. Then, problem \eqref{eqn:original_Kmeans} is 
equivalent to finding a discrete probability measure $G$ which has finite 
number of support points and solves:
\vspace{-6pt}
\begin{eqnarray}
\mathop {\inf }\limits_{G \in \mathcal{O}_{k}(\Theta)}{W_{2}^{2}(G,P_{n})}. \label{eqn:Wasserstein_K_means}
\end{eqnarray} 
Due to the inclusion of Wasserstein metric in its formulation, we call this
a \emph{Wasserstein means problem}. This problem can be further thought of as a 
Wasserstein barycenter problem where $N=1$. In light of this observation, as noted by
\citep{Cuturi-2014}, the algorithm for finding the Wasserstein barycenter offers an 
alternative for the popular Loyd's algorithm for determing local minimum of the K-means objective.

%% file: arxiv_nested_multilevel.tex
Given $m$ groups of $n_{j}$ exchangeable data points $X_{j,i}$ where $1 
\leq j \leq m, 1 \leq i \leq n_{j}$, i.e., data are presented in a two-level grouping structure, 
our goal is to learn about the two-level clustering structure of the data. We want
to obtain simultaneously local clusters for each data group, and global clusters among all groups. 

\subsection{Multilevel Wasserstein Means (MWM) Algorithm} \label{Section:multilevel_kmeans}
For any $j=1,\ldots, m$, we denote the empirical measure for group $j$ by $P_{n_{j}}
^{j}:=\dfrac{1}{n_{j}}\sum \limits_{i=1}^{n_{j}}{\delta_{X_{j,i}}}$. Throughout 
this section, for simplicity of exposition we assume that the number of 
both local and global clusters are either known or bounded above by a given number. In 
particular, for local clustering we allow group $j$ to have at most $k_{j}$ clusters 
for $j=1,\ldots, m$. For global clustering, we assume to have $M$ group (Wasserstein) means
among the $m$ given groups.

\paragraph{High level idea}
For local clustering, for each $j = 1,\ldots, m$,
performing a K-means clustering for group $j$, as expressed by
\eqref{eqn:Wasserstein_K_means}, can be viewed as finding a finite discrete measure $G_{j} 
\in \mathcal{O}_{k_{j}}(\Theta)$ that minimizes squared Wasserstein distance $W_{2}^{2}
(G_{j},P_{n_{j}}^{j})$. For global clustering, we are interested in 
obtaining clusters out of $m$ groups, each of which is now represented by the
discrete measure $G_j$, for $j=1,\ldots,m$. 
Adopting again the viewpoint of Eq.~\eqref{eqn:Wasserstein_K_means}, 
provided that all of $G_{j}$s are given, we can apply $K$-means quantization method 
to find their distributional clusters. The global clustering in the
space of measures of measures on $\Theta$ can be succintly expressed by
\vspace{-6pt}
\begin{eqnarray}
\mathop {\inf }\limits_{\mathcal{H} \in \mathcal{E}_{M}(\mathcal{P}_{2}(\Theta))}{W_{2}^{2}\biggr(
\mathcal{H},\dfrac{1}{m}\sum \limits_{j=1}^{m}{\delta_{G_{j}}}\biggr)}. \nonumber
\end{eqnarray}
However, $G_{j}$ are not known --- they have to be optimized through local clustering in 
each data group.
\paragraph{MWM problem formulation} We have arrived at an objective function for jointly
optimizing over both local and global clusters
\vspace{-6pt}
\begin{eqnarray}
\mathop {\inf }\limits_{\substack {G_{j} \in \mathcal{O}_{k_{j}}(\Theta), \\\mathcal{H} \in 
\mathcal{E}_{M}(\mathcal{P}_{2}(\Theta))}}{\mathop {\sum }\limits_{j=1}^{m}{W_{2}
^{2}(G_{j},P_{n_{j}}^{j})}}
+ W_{2}^{2}(\mathcal{H},\dfrac{1}{m}\mathop {\sum }\limits_{j=1}^{m}{\delta_{G_{j}}}). \label{eqn:multilevel_Kmeans_typeone}
\end{eqnarray}

We call the above optimization the problem of \emph{Multilevel Wasserstein Means (MWM)}. 
The notable feature of MWM is that its loss function consists of two types of distances 
associated with the hierarchical data structure:
one is distance in the space of measures, e.g., $W_{2}^{2}(G_{j},P_{n_{j}}^{j})$, 
and the other in space of measures of measures, e.g., 
$W_{2}^{2}(\mathcal{H},\dfrac{1}{m}\mathop {\sum }\limits_{j=1}^{m}{\delta_{G_{j}}})$. By 
adopting K-means optimization to both local and global clustering, the multilevel Wasserstein means 
problem might look formidable at the first sight. 
Fortunately, it is possible to simplify this original formulation substantially,
by exploiting the structure of $\Hcal$.

Indeed, we can show that formulation \eqref{eqn:multilevel_Kmeans_typeone} is
equivalent to the following optimization problem, which looks much simpler as
it involves only measures on $\Theta$:
\begin{eqnarray}
\mathop {\inf }\limits_{G_{j} \in \mathcal{O}_{k_{j}}(\Theta), \Hbold}
{\mathop {\sum }\limits_{j=1}^{m}{W_{2}^{2}(G_{j},P_{n_{j}}^{j})}+\dfrac{d_{W_{2}}^{2}(G_{j},\Hbold)}{m}} \label{eqn:multilevel_K_means_typeone_first}
\end{eqnarray}
where $d_{W_{2}}^{2}(G,\Hbold) := \mathop {\min } \limits_{1 \leq i \leq M}{W_{2}^{2}(G,H_{i})}$ and $\Hbold=(H_{1},\ldots,H_{M})$,
with each $H_i \in \mathcal{P}_2(\Theta)$. The proof of this 
equivalence is deferred to Proposition \ref{lemma:equivalence_multilevels_Kmeans} in the Supplement. 
Before going into to the details of the algorithm for solving 
\eqref{eqn:multilevel_K_means_typeone_first}
in Section \ref{Section:mwm_algorithm}, we shall present some simpler cases,
which help to illustrate some properties of the optimal solutions of 
\eqref{eqn:multilevel_K_means_typeone_first},
while providing insights of subsequent developments of the MWM formulation.
Readers may proceed directly to Section \ref{Section:mwm_algorithm} 
for the description of the algorithm in the first reading.
\subsubsection{Properties of MWM in special cases} \label{Section:mwm_specical_cases}
\paragraph{Example 1.} Suppose $k_{j}=1$ and $n_{j}=n$ for all $1 \leq j \leq m$, and $M=1$. Write
$\Hbold = H \in \mathcal{P}_2(\Theta)$. Under 
this setting, the objective function \eqref{eqn:multilevel_K_means_typeone_first} can be 
rewritten as
\begin{eqnarray}
\mathop {\inf }\limits_{\substack {\theta_{j} \in \Theta, \\ H \in \mathcal{P}_{2}(\Theta)}}{\sum \limits_{j=1}^{m}{\sum \limits_{i=1}^{n}{\|\theta_{j}-X_{j,i}\|^{2}}}}
+W_{2}^{2}(\delta_{\theta_{j}},H)/m, \label{eqn:special_case_multilevel_Kmeans_one} 
\end{eqnarray}
where $G_{j}=\delta_{\theta_{j}}$ for any $1 \leq j \leq m$. From the result of Theorem 
A.1 in the Supplement, 
\vspace{-6pt}
\begin{eqnarray}
\mathop {\inf } \limits_{\theta_{j} \in \Theta}{\sum \limits_{j=1}^{m}{W_{2}^{2}(\delta_{\theta_{j}},H)}} & \geq & \mathop {\inf }\limits_{H \in \mathcal{E}_{1}(\Theta)}{\sum \limits_{j=1}^{m}{W_{2}^{2}(G_{j},H)}} \nonumber \\
& = & \sum \limits_{j=1}^{m}{\|\theta_{j}-(\sum \limits_{i=1}^{m}{\theta_{i}})/m\|^{2}}, \nonumber
\end{eqnarray}
where second infimum is achieved when $H=\delta_{(\sum \limits_{j=1}^{m}{\theta_{j}})/m}$. 
Thus, objective function \eqref{eqn:special_case_multilevel_Kmeans_one} may
be rewritten as
\vspace{-6pt}
\begin{eqnarray}
\mathop {\inf }\limits_{\theta_{j} \in \Theta}{\sum \limits_{j=1}^{m}{\sum \limits_{i=1}^{n}{\|\theta_{j}-X_{j,i}\|^{2}}}} +\| m\theta_{j}-(\sum \limits_{l=1}^{m}{\theta_{l}})\|^{2}/m^{3}. \nonumber
\end{eqnarray}
Write $\overline{X}_{j}=(\sum \limits_{i=1}^{n}{X_{j,i}})/n$ for all $1 \leq j \leq m$. 
As $m \geq 2$, we can check that the unique optimal solutions for the above optimization 
problem are $\theta_{j}=\biggr((m^2n+1)\overline{X}_{j}+\sum \limits_{i \neq j}
{\overline{X}_{i}}\biggr)/(m^{2}n+m)$ for any $1 \leq j \leq m$. If we further assume that 
our data $X_{j,i}$ are i.i.d samples from probability measure $P^{j}$ having mean $\mu_{j}
=E_{X \sim P^{j}}(X)$ for any $1 \leq j \leq m$, the previous result implies that $\theta_{i} 
\not \to \theta_{j}$ for almost surely as long as $\mu_{i} \neq \mu_{j}$. As a 
consequence, if $\mu_{j}$ are pairwise different, the multi-level Wasserstein means under 
that simple scenario of \eqref{eqn:multilevel_K_means_typeone_first} will not have identical 
centers among local groups. 

On the other hand, we have $W_{2}^{2}(G_{i},G_{j})=\|\theta_{i}-\theta_{j}\|^{2}=
\biggr(\dfrac{mn}{mn+1}\biggr)^{2}\|\overline{X}_{i}-\overline{X}_{j}\|^{2}$. Now, 
from the definition of Wasserstein distance
\vspace{-6pt}
\begin{eqnarray}
W_{2}^{2}(P_{n}^{i},P_{n}^{j}) & = & \mathop {\min }\limits_{\sigma}{\dfrac{1}{n}\sum \limits_{l=1}^{n}{\|X_{i,l}-X_{j,\sigma(l)}\|^{2}}} \nonumber \\
& \geq & \|\overline{X}_{i}-\overline{X}_{j}\|^{2}, \nonumber
\end{eqnarray}
where $\sigma$ in the above sum varies over all the permutation of $\left\{1,2,\ldots,n
\right\}$ and the second inequality is due to Cauchy-Schwarz's inequality. It implies that as 
long as $W_{2}^{2}(P_{n}^{i},P_{n}^{j})$ is small, the optimal solution $G_{i}$ and $G_{j}
$ of \eqref{eqn:special_case_multilevel_Kmeans_one} will be sufficiently close to each 
other. By letting $n \to \infty$, we also achieve the same conclusion regarding the 
asymptotic behavior of $G_{i}$ and $G_{j}$ with respect to $W_2(P^{i},P^{j})$.

\paragraph{Example 2.} $k_{j}=1$ and $n_{j}=n$ for all $1 \leq  j \leq m$ and $M=2$. 
Write $\Hbold = (H_1,H_2)$.
Moreover, assume that there is a strict subset A of $\left\{1,2,\ldots,m\right\}$ 
such that 
\vspace{-6pt}
\begin{eqnarray}
& & \mathop {\max }\biggr\{\mathop {\max }\limits_{i, j \in A}{W_{2}(P_{n}^{i},P_{n}^{j})}, \nonumber \\
& & \mathop {\max }\limits_{i, j \in A^{c}}{W_{2}(P_{n}^{i},P_{n}^{j})}\biggr\} \ll \mathop {\min }\limits_{i \in A, j \in A^{c}}{W_{2}(P_{n}^{i},P_{n}^{j})}, \nonumber 
\end{eqnarray}
i.e., the distances of empirical measures $P_{n}^{i}$ and $P_{n}^{j}$ when $i$ and $j$ 
belong to the same set $A$ or $A^{c}$ are much less than those when $i$ and $j$ do not 
belong to the same set. Under this condition, by using the argument from part (i) we can write 
the objective function \eqref{eqn:multilevel_K_means_typeone_first} as 
\vspace{-6pt}
\begin{eqnarray}
\mathop {\inf }\limits_{\substack {\theta_{j} \in \Theta, \\ H_{1} \in \mathcal{P}_{2}(\Theta)}}{\sum \limits_{j \in A}{\sum \limits_{i=1}^{n}{\|\theta_{j}-X_{j,i}\|^{2}}}+\dfrac{W_{2}^{2}(\delta_{\theta_{j}},H_{1})}{|A|}}+ \nonumber \\ 
\mathop {\inf }\limits_{\substack {\theta_{j} \in \Theta, \\ H_{2} \in \mathcal{P}_{2}(\Theta)}}{\sum \limits_{j \in A^{c}}{\sum \limits_{i=1}^{n}{\|\theta_{j}-X_{j,i}\|^{2}}}+\dfrac{W_{2}^{2}(\delta_{\theta_{j}},H_{2})}{|A^{c}|}}. \nonumber
\end{eqnarray}
The above objective function suggests that the optimal solutions $\theta_{i}$, $\theta_{j}$ 
(equivalently, $G_{i}$ and $G_{j}$) will not be close to each other as long as $i$ and 
$j$ do not belong to the same set $A$ or $A^{c}$, i.e., $P_{n}^{i}$ and $P_{n}^{j}$ are 
very far. Therefore, the two groups of ``local'' measures $G_{j}$ do not share atoms under that 
setting of empirical measures.

The examples examined above indicate that the MWM problem in general
do not ``encourage'' the local measures $G_{j}$ to share atoms among each other in its solution. Additionally, 
when the empirical measures of local groups are very close, it may also suggest that they 
belong to the same cluster and the distances among optimal local measures $G_{j}$ can be 
very small. 

\subsubsection{Algorithm Description} \label{Section:mwm_algorithm}
Now we are ready to describe our algorithm in the general case. This is
a procedure for finding a local minimum of Problem \eqref{eqn:multilevel_K_means_typeone_first} and
is summarized in Algorithm \ref{alg:multilevels_Wasserstein_means}. 
\begin{algorithm}[tb]
   \caption{Multilevel Wasserstein Means (MWM)}
   \label{alg:multilevels_Wasserstein_means}
\begin{algorithmic}
   \STATE {\bfseries Input:} Data $X_{j,i}$, Parameters $k_{j}$, $M$.
   \STATE {\bfseries Output:} prob. measures $G_{j}$ and elements $H_{i}$ of $\Hbold$.
   \STATE Initialize measures $G_{j}^{(0)}$, elements $H_{i}^{(0)}$ of $\Hbold^{(0)}$, $t=0$.
   \WHILE{$Y_{j}^{(t)}, b_{j}^{(t)}, H_{i}^{(t)}$ have not converged}
   \STATE 1. Update $Y_{j}^{(t)}$ and $b_{j}^{(t)}$ for $1 \leq j \leq m$:
   \FOR{$j=1$ {\bfseries to} $m$}
   \STATE $i_{j} \leftarrow \mathop {\arg \min}\limits_{1 \leq u \leq M}{W_{2}^{2}(G_{j}^{(t)},H_{u}^{(t)})}$.
   \STATE $G_{j}^{(t+1)} \leftarrow \mathop {\arg \min }\limits_{G_{j} \in \mathcal{O}_{k_{j}}(\Theta)}{W_{2}^{2}(G_{j},P_{n_{j}}^{j})}+$\\$+W_{2}^{2}(G_{j},H_{i_{j}}^{(t)})/m$.
   \ENDFOR
   \STATE 2. Update $H_{i}^{(t)}$ for $1 \leq i \leq M$:
   \FOR{$j=1$ {\bfseries to} $m$}
   \STATE $i_{j} \leftarrow \mathop {\arg \min}\limits_{1 \leq u \leq M}{W_{2}^{2}(G_{j}^{(t+1)},H_{u}^{(t)})}$.
   \ENDFOR
   \FOR{$i=1$ {\bfseries to} $M$}
   \STATE $C_{i} \leftarrow \left\{l: i_{l}=i\right\}$ for $1 \leq i \leq M$.
   \STATE $H_{i}^{(t+1)} \leftarrow \mathop {\arg \min }\limits_{H_{i} \in \mathcal{P}_{2}(\Theta)}{\sum \limits_{l \in C_{i}}{W_{2}^{2}(H_{i}, G_{l}^{(t+1)})}}$.
   \ENDFOR
   \STATE 3. $t \leftarrow t+1$.
   \ENDWHILE
\end{algorithmic}
\end{algorithm}
We prepare the following details regarding the initialization and updating steps required by
the algorithm: 
\begin{itemize}
\item The initialization of local measures $G_{j}^{(0)}$ (i.e., the initialization of their atoms and weights) can be 
obtained by performing $K$-means clustering on local data $X_{j,i}$ for $1 \leq j \leq m$.
The initialization of elements $H_{i}^{(0)}$ of $H^{(0)}$ is based on 
a simple extension of the K-means algorithm. Details are given in Algorithm \ref{alg:three_stages_K_means} in the Supplement;
\item The updates $G_{j}
^{(t+1)}$ can be computed efficiently by simply using algorithms from \cite{Cuturi-2014} 
to search for local solutions of these barycenter problems within the space $\mathcal{O}
_{k_{j}}(\Theta)$ from the atoms and weights of 
$G_{j}^{(t)}$; 
\item Since all $G_{j}^{(t+1)}$ are finite discrete 
measures, finding the updates for $H_{i}^{(t+1)}$ over the whole space $\mathcal{P}_{2}
(\Theta)$ can be reduced to searching for a local solution within space $\mathcal{O}
_{l^{(t)}}$ where $l^{(t)}=\sum \limits_{j \in C_{i}}{|\text{supp}(G_{j}^{(t+1)})|}-|C_{i}|$ from the global atoms $H_{i}^{(t)}$ of $\Hbold^{(t)}$
(Justification of this reduction is derived from Theorem \ref{theorem:upperbound_barycenter} in the Supplement). 
This again can be done by utilizing algorithms from \cite{Cuturi-2014}. Note that, as $l^{(t)}$ becomes very large when $m$ is large, to speed up the computation of Algorithm 
\ref{alg:multilevels_Wasserstein_means} we impose a threshold $L$, e.g., $L=10$, for 
$l^{(t)}$ in its implementation. 
\end{itemize}
The following guarantee for Algorithm~\ref{alg:multilevels_Wasserstein_means}
can be established:
\begin{theorem}\label{theorem:local_convergence_multilevel_Kmeans}
Algorithm \ref{alg:multilevels_Wasserstein_means} monotonically decreases the objective 
function \eqref{eqn:multilevel_Kmeans_typeone} of the MWM formulation.
\end{theorem}
\subsection{Multilevel Wasserstein Means with Sharing} \label{Section:constraint_multilevels_Kmeans}
As we have observed from the analysis of several specific cases,
the {\bf multilevel Waserstein means} formulation
may not encourage the sharing components locally among $m$ groups in its solution.
However, enforced sharing has been demonstrated to be a very useful technique, which
leads to the ``borrowing of strength'' among different parts of the model, consequentially
improving the inferential efficiency~\citep{Teh-etal-06,Nguyen-2016}. In this section, 
we seek to encourage the borrowing of strength among groups by imposing additional
constraints on the atoms of $G_{1},\ldots,G_{m}$ in the original MWM
formulation \eqref{eqn:multilevel_Kmeans_typeone}. Denote $\mathcal{A}_{M,
\mathcal{S}_{K}}=\biggr\{G_{j} \in \mathcal{O}_{K}(\Theta), \ \Hcal \in \mathcal{E}_{M}
(\mathcal{P}(\Theta)): \text{supp}(G_{j}) \subseteq \mathcal{S}_{K}\ \forall 1 \leq j \leq 
m \biggr\}$
for any given $K, M \geq 1$ where the constraint set $\mathcal{S}_{K}$ has exactly $K$ 
elements. To simplify the exposition, let us assume that $k_{j}=K$ for all $1 
\leq j \leq m$. Consider the following locally constrained version of the
multilevel Wasserstein means problem
\begin{eqnarray}
\mathop {\inf }{\mathop {\sum }\limits_{j=1}^{m}{W_{2}^{2}(G_{j},P_{n_{j}}^{j})}}
+W_{2}^{2}(\Hcal,\dfrac{1}{m}\mathop {\sum }\limits_{j=1}^{m}{\delta_{G_{j}}}). \label{eqn:local_constraint_multilevels_Kmeans_typeone}
\end{eqnarray}
where $\mathcal{S}_{K}, \ G_{j},\Hcal \in \mathcal{A}_{M,\mathcal{S}_{K}}$ in the above infimum. We call the above optimization the problem of \emph{Multilevel Wasserstein Means with Sharing (MWMS)}. The 
local constraint assumption $\text{supp}(G_{j})\subseteq \mathcal{S}_{K}$ had been 
utilized previously in the literature --- see for example the work of \citep{Kulis-2012}, 
who developed an optimization-based approach to the inference of the HDP~\citep{Teh-etal-06},
which also encourages explicitly the sharing of local group means among local clusters. 
Now, we can rewrite objective function 
\eqref{eqn:local_constraint_multilevels_Kmeans_typeone} as follows
\vspace{-6pt}
\begin{eqnarray}
\mathop {\inf }\limits_{\mathcal{S}_{K},G_{j} , \Hbold \in \mathcal{B}_{M,\mathcal{S}_{K}}}{\mathop {\sum }\limits_{j=1}^{m}{W_{2}^{2}(G_{j},P_{n_{j}}^{j})}+\dfrac{d_{W_{2}}^{2}(G_{j},\Hbold)}{m}} \label{eqn:local_constraint_multilevels_Kmeans_typetwo}
\end{eqnarray}
where  $\mathcal{B}_{M,\mathcal{S}_{K}}=\biggr\{G_{j} \in \mathcal{O}_{K}(\Theta), \ \Hbold=(H_{1},\ldots,H_{M}): 
\text{supp}(G_{j})\subseteq \mathcal{S}_{K}\ \forall 1 \leq j \leq m \biggr\}$. 
The high level idea of finding local minimums of objective function \eqref{eqn:local_constraint_multilevels_Kmeans_typetwo} 
is to first, update the elements of constraint set $\mathcal{S}_{K}$ to provide the 
supports for local measures $G_{j}$ and then, obtain the weights of these measures as 
well as the elements of global set $H$ by computing appropriate Wasserstein barycenters. 
Due to space constraint, the details of these steps of the MWMS Algorithm (Algorithm \ref{alg:local_constraint_multilevels_Wasserstein_means})
are deferred to the Supplement.

%% file: arxiv_nested_consistency.tex
We proceed to establish consistency for the estimators introduced in the previous
section. For the brevity of the presentation, we only focus on the MWM method;
consistency for MWMS can be obtained in a similar fashion. 
Fix $m$, and assume that $P^{j}$ is the 
true distribution of data $X_{j,i}$ for $j = 1,\ldots,m$. Write
$\vec{G}=(G_{1},\ldots,G_{m})$ and $\vec{n}=(n_{1},\ldots,n_{m})$. We say 
$\vec{n} \to \infty$ if $n_{j} \to \infty$ for $j=1,\ldots, m$. Define the following 
functions
\vspace{-6pt}
\begin{eqnarray}
f_{\vec{n}}(\vec{G},\Hcal)=\mathop {\sum }\limits_{j=1}^{m}{W_{2}^{2}(G_{j},P_{n_{j}}^{j})}+W_{2}^{2}(\Hcal,\dfrac{1}{m}\mathop {\sum }\limits_{j=1}^{m}{\delta_{G_{j}}}), \nonumber \\
f(\vec{G},\Hcal)=\mathop {\sum }\limits_{j=1}^{m}{W_{2}^{2}(G_{j},P^{j})}+W_{2}^{2}(\Hcal,\dfrac{1}{m}\mathop {\sum }\limits_{j=1}^{m}{\delta_{G_{j}}}), \nonumber
\end{eqnarray}
where $G_{j} \in \mathcal{O}_{k_{j}}(\Theta)$, $\Hcal \in \mathcal{E}_{M}(\mathcal{P}(\Theta))$ as $1 \leq j \leq m$. 
The first consistency property of the WMW formulation:
\begin{theorem} \label{theorem:objective_consistency_multilevel_Wasserstein_means} 
Given that $P^{j} \in \mathcal{P}_{2}(\Theta)$ for $1 \leq j \leq m$. Then,
there holds almost surely, as $\vec{n} \to \infty$
\vspace{-6pt}
\begin{eqnarray}
\mathop {\inf }\limits_{\substack {G_{j} \in \mathcal{O}_{k_{j}}(\Theta), \\ \Hcal \in \mathcal{E}_{M}(\mathcal{P}_{2}(\Theta))}}f_{\vec{n}}(\vec{G},\Hcal) - \mathop {\inf }\limits_{\substack {G_{j} \in \mathcal{O}_{k_{j}}(\Theta), \\ \Hcal \in \mathcal{E}_{M}(\mathcal{P}_{2}(\Theta))}}f(\vec{G},\Hcal) \rightarrow 0. \nonumber
\end{eqnarray}

\end{theorem}
The next theorem establishes that the ``true'' global and local clusters can be
recovered. To this end, assume that for each $\vec{n}$ there is an optimal solution $
(\widehat{G}_{1}^{n_{1}},\ldots,\widehat{G}_{m}^{n_{m}},\widehat{\Hcal}^{\vec{n}})$ or in 
short $(\vec{\widehat{G}}^{\vec{n}},\Hcal^{\vec{n}})$ of the objective function 
\eqref{eqn:multilevel_Kmeans_typeone}. Moreover, there exist a (not necessarily unique)
optimal solution minimizing $f(\vec{G},\Hcal)$ over $G_{j} \in \mathcal{O}_{k_{j}}(\Theta)$ and $\Hcal \in 
\mathcal{E}_{M}(\mathcal{P}_{2}(\Theta))$. Let $\mathcal{F}$ be the collection 
of such optimal solutions. For any $G_{j} \in \mathcal{O}_{k_{j}}(\Theta)$ and $\Hcal \in 
\mathcal{E}_{M}(\mathcal{P}_{2}(\Theta))$, define
\vspace{-6pt}
\begin{eqnarray}
d(\vec{G},\Hcal,\mathcal{F})=\inf \limits_{(\vec{G}^{0}, \Hcal^{0}) \in \mathcal{F}}\sum \limits_{j=1}^{m}{W_{2}^{2}(G_{j},G_{j}^{0})} \nonumber
+W_{2}^{2}(\Hcal,\Hcal^{0}). \nonumber
\end{eqnarray}
Given the above assumptions, we have the following result regarding the convergence of $(\widehat{\vec{G}}^{\vec{n}},\Hcal^{\vec{n}})$:
\begin{theorem} \label{theorem:convergence_measures_multilevel_Wasserstein_means}
Assume that $\Theta$ is bounded and $P^{j} \in \mathcal{P}_{2}(\Theta)$ for all $1 \leq j 
\leq m$. Then, we have $d(\vec{\widehat{G}}^{\vec{n}},\widehat{\Hcal}
^{\vec{n}},\mathcal{F}) \to 0$ as $\vec{n} \to \infty$ almost surely.
\end{theorem}
\paragraph{Remark:}(i) The assumption $\Theta$ is bounded is just for the convenience of 
proof argument. We believe that the conclusion of this theorem may still hold when $\Theta 
= \mathbb{R}^{d}$. (ii) If $|\mathcal{F}|=1$, i.e., there exists an unique optimal solution $
\vec{G}^{0},\Hcal^{0}$ minimizing $f(\vec{G},\Hcal)$ over $G_{j} \in \mathcal{O}_{k_{j}}(\Theta)
$ and $\Hcal \in \mathcal{E}_{M}(\mathcal{P}_{2}(\Theta))$, the result of Theorem 
\ref{theorem:convergence_measures_multilevel_Wasserstein_means} implies that $W_{2}
(\widehat{G}_{j}^{n_{j}},G_{j}^{0}) \to 0$ for $1 \leq j \leq m$ and $W_{2}(\widehat{\Hcal}
^{\vec{n}},\Hcal^{0}) \to 0$ as $\vec{n} \to \infty$.

\begin{figure*}[ht]
\centerline{\includegraphics[width=0.9\textwidth]{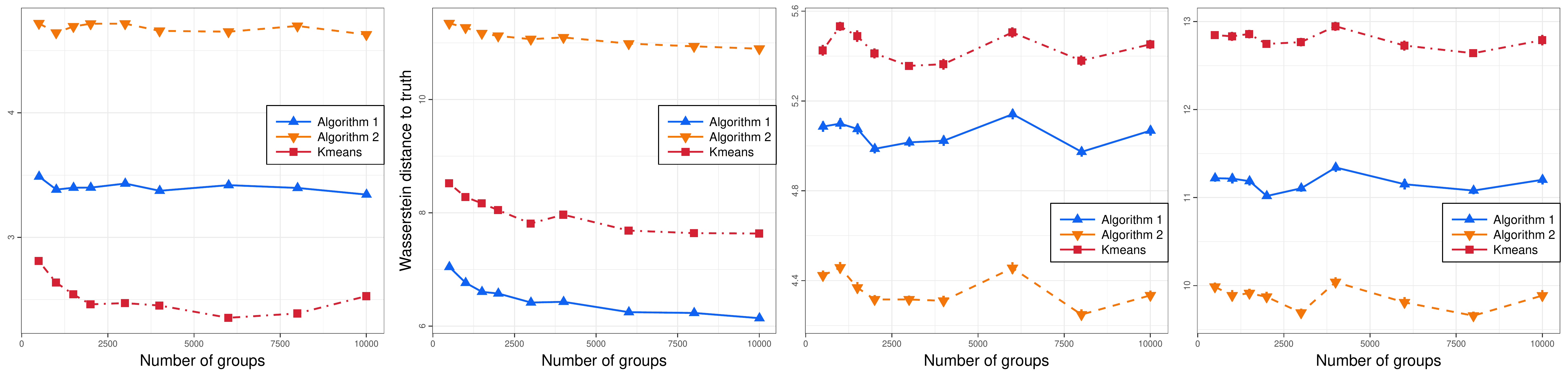}}
\caption{Data with a lot of small groups: (a) NC data with constant variance; (b) NC data with non-constant variance; (c)
LC data with constant variance; (d) LC data with non-constant variance}
\label{fig:simul_M}
\end{figure*}

\begin{figure*}[ht]
\centerline{\includegraphics[width=0.9 \textwidth]{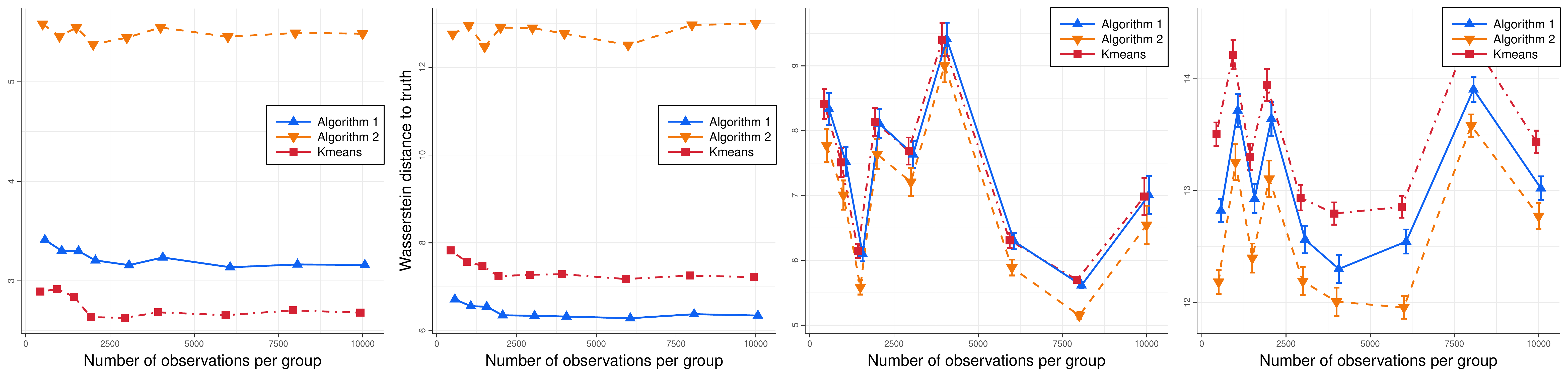}}
\caption{Data with few big groups: (a) NC data with constant variance; (b) NC data with non-constant variance; (c) LC
data with constant variance; (d) LC data with non-constant variance}
\label{fig:simul_N}
\end{figure*}

%% file: arxiv_nested_simulation.tex
\subsection{Synthetic data}
In this section, we are interested in evaluating the effectiveness of both
MWM and MWMS clustering algorithms by considering different synthetic data generating processes. 
Unless otherwise specified, we set the number of groups $m=50$, number of observations 
per group $n_{j}=50$ in $d=10$ dimensions, number of global clusters $M=5$ with 6 atoms. 
For Algorithm \ref{alg:multilevels_Wasserstein_means} (MWM)
local measures $G_{j}$ have 5 atoms each; for Algorithm \ref{alg:local_constraint_multilevels_Wasserstein_means} (MWMS) number of atoms in constraint set $S_K$ is 50. 
As a benchmark for the comparison we will use a basic 3-stage K-means approach
(the details of which can be found in the Supplement).
The Wasserstein distance between the estimated distributions (i.e. $\hat 
G_1,\ldots,\hat G_m$; $\hat H_1,\ldots,\hat H_M$) and the data generating ones will
be used as the comparison metric. 

Recall that the MWM formulation does not impose constraints on the atoms of $G_{i}$,
while the MWMS formulation explicitly enforces the sharing of atoms
across these measures.
We used multiple layers of mixtures while adding Gaussian noise at each layer to generate global and local 
clusters and the no-constraint (NC) data. We varied number of groups $m$ 
from 500 to 10000. We notice that the 3-stage K-means algorithm performs the best 
when there is no constraint structure \emph{and} variance is constant across clusters (Fig. 
\ref{fig:simul_M}(a) and \ref{fig:simul_N}(a)) --- this is, not surprisingly, a favorable setting for the basic K-means method.
As soon as we depart from the (unrealistic) constant-variance, no-sharing assumption, both of our
algorithms start to outperform the basic three-stage K-means.
The superior performance is most pronounced with local-constraint (LC) data (with or without constant variance conditions). 
See Fig. \ref{fig:simul_M}(c,d).
It is worth noting that even when group variances are constant, the 3-stage K-means is no longer
longer effective because now fails to account for the shared structure. 
When $m=50$ and group sizes are larger, we set $S_K=15$. Results are reported in 
Fig. \ref{fig:simul_N} (c), (d). 
These results demonstrate the effectiveness and flexibility of our both algorithms.


\subsection{Real data analysis} 
\begin{figure*}[t]
\begin{centering}
\subfloat[\label{fig:labelme_clusters}]{\begin{centering}
\includegraphics[width=0.2\paperwidth]{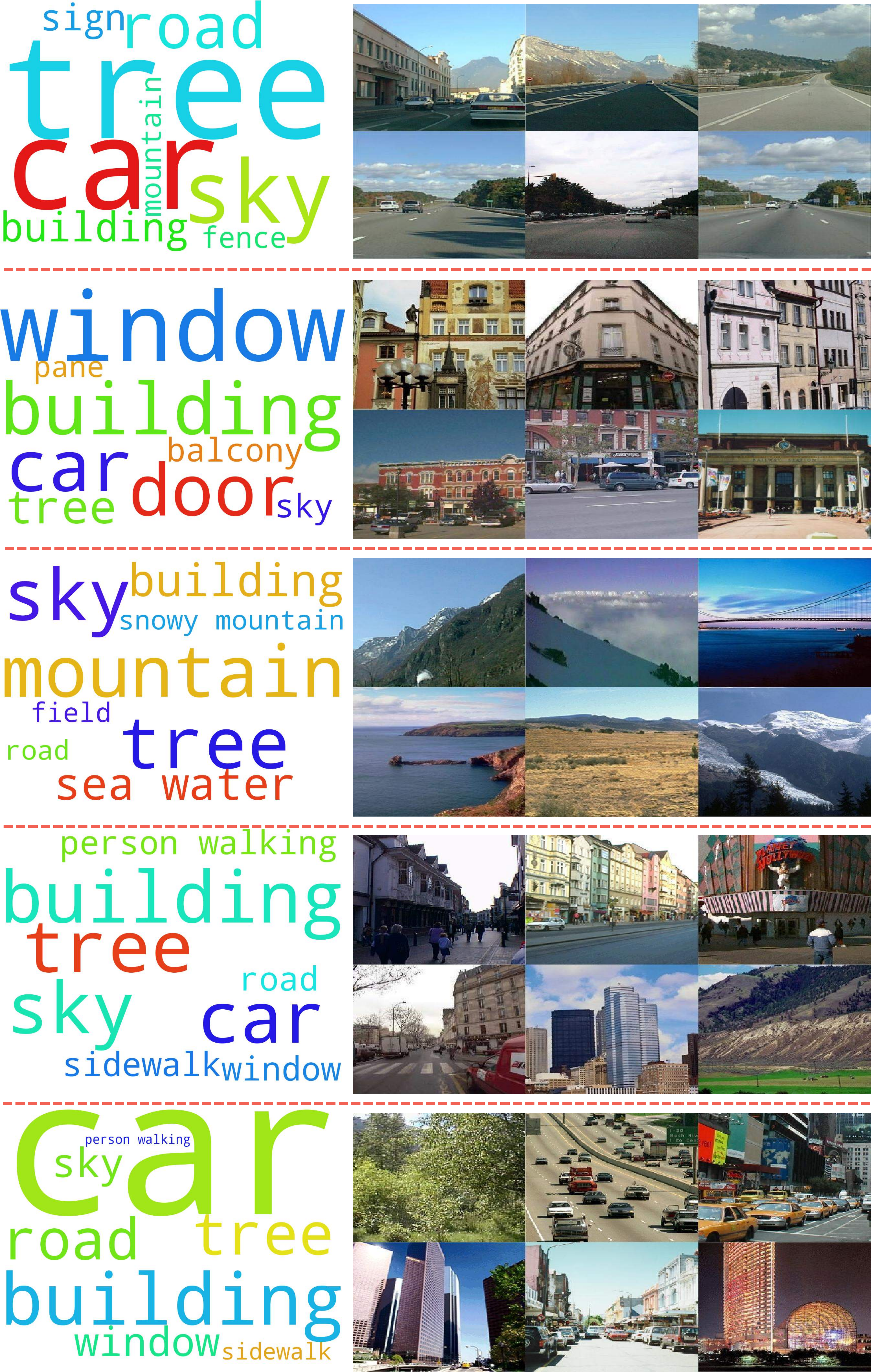}
\par\end{centering}

}\qquad{}\subfloat[
\label{fig:SL-graph}
]{\begin{centering}
\includegraphics[width=0.3\paperwidth]{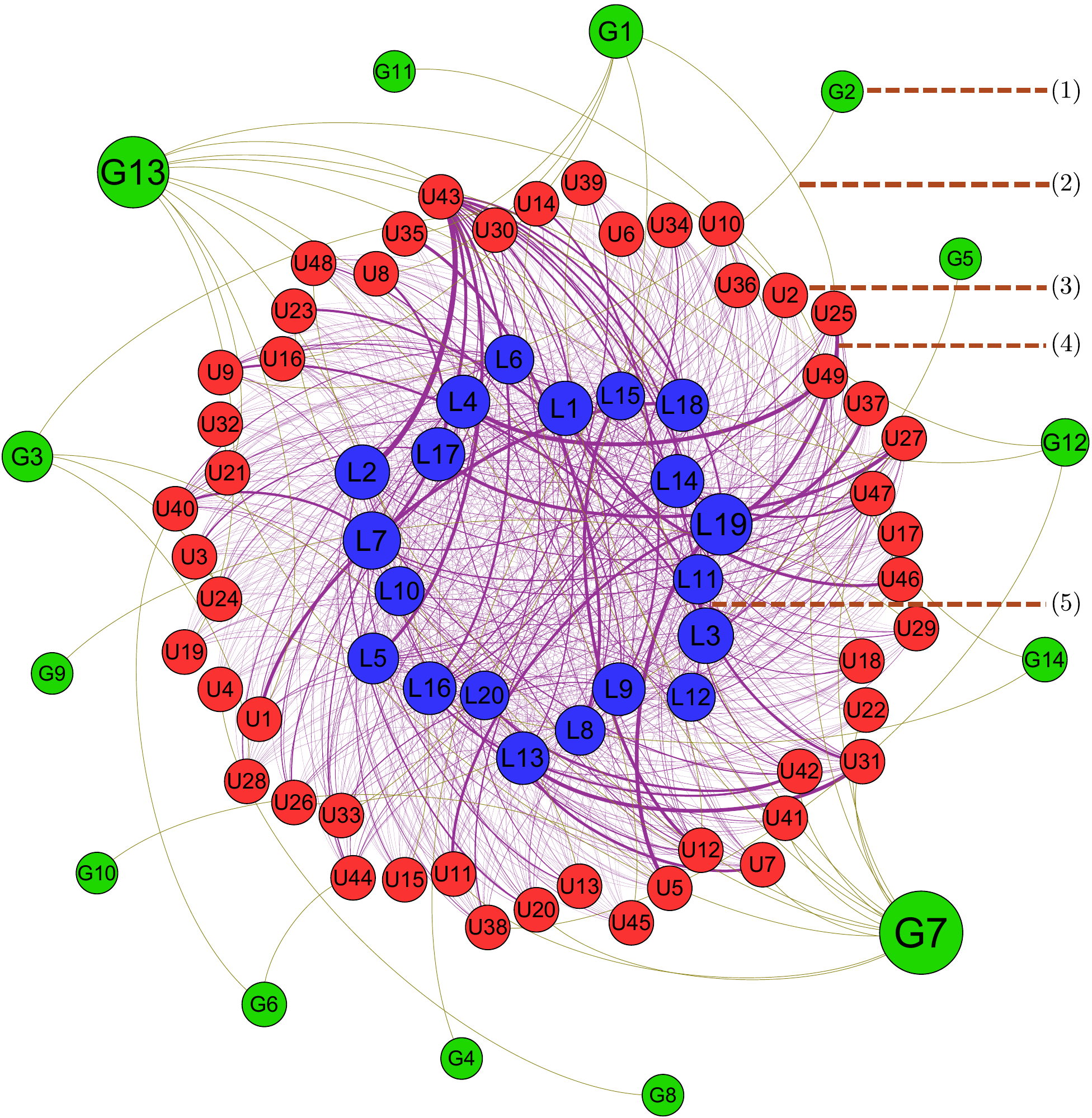}
\par\end{centering}

}
\par\end{centering}

\caption{
Clustering representation for two datasets: (a) Five image clusters
from \emph{Labelme} data discovered by MWMS algorithm: tag-clouds
on the left are accumulated from all images in the clusters while
six images on the right are randomly chosen images in that cluster;
(b) StudentLife discovered network with three node groups: (1) discovered
student clusters, (3) student nodes, (5) discovered activity location
(from Wifi data); and two edge groups: (2) Student to cluster assignment,
(4) Student involved to activity location. Node sizes (of discovered
nodes) depict the number of element in clusters while edge sizes between
\emph{Student} and \emph{activity location }represent the popularity
of student's activities.
}
\end{figure*}

We applied our multilevel clustering algorithms to two real-world datasets: LabelMe and StudentLife.

\begin{figure}
\centering{}\includegraphics[width=0.17\textwidth]{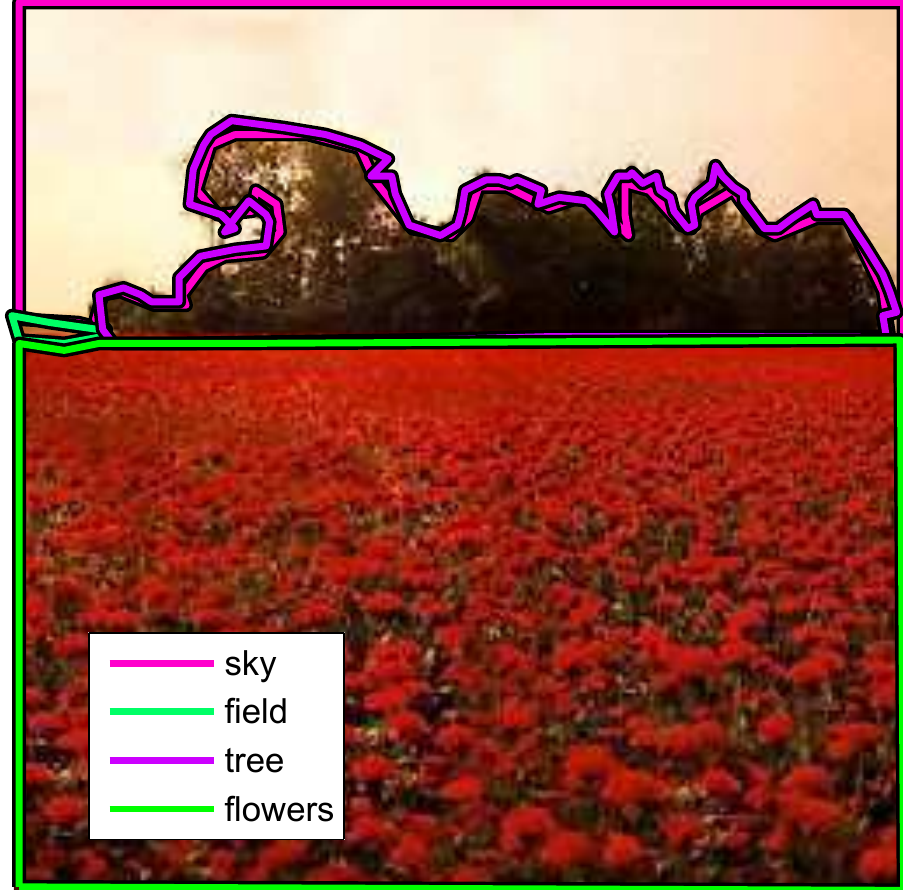}\qquad{}\qquad{}\includegraphics[width=0.22\textwidth]{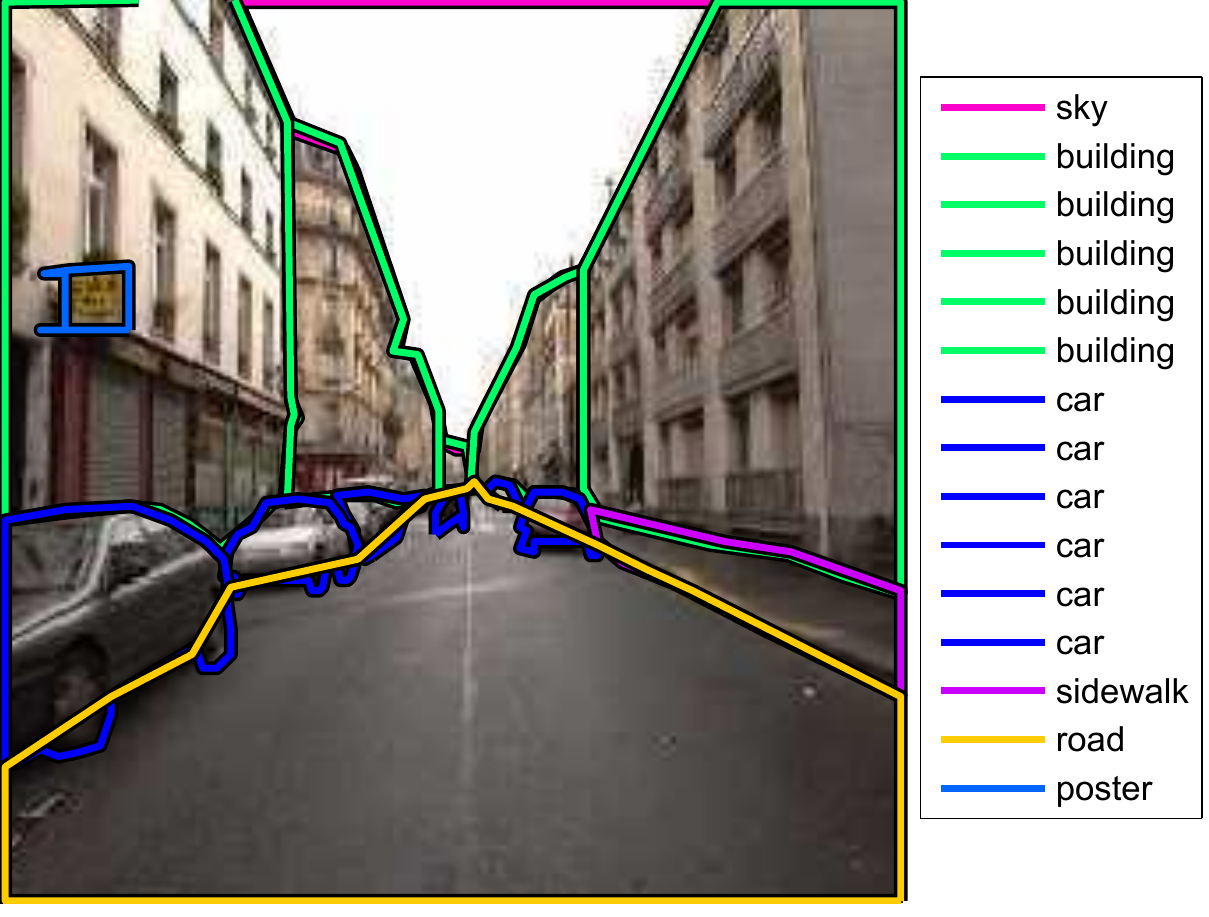}\caption{Examples of images used in LabelMe dataset. Each image consists of
different annotated regions.\label{fig: LabelMeExample}}
\end{figure}

\textbf{LabelMe dataset} consists of $2,688$ annotated
images which are classified into 8 scene categories including \emph{tall
buildings, inside city, street, highway, coast, open country, mountain,}
and \emph{forest} \cite{Oliva-2001} . Each image contains multiple
annotated regions. Each region, which is annotated by users, represents
an object in the image.  As shown in Figure \ref{fig: LabelMeExample}, the left
image is an image from \emph{open country }category and contains
4 regions while the right panel denotes an image of \emph{tall buildings}
category including 16 regions. Note that the regions in each image can
be overlapped. We remove the images containing less then 4 regions
and obtain $1,800$ images. 

We then extract GIST feature \cite{Oliva-2001} for each region in a image.
GIST is a visual descriptor to represent perceptual dimensions and
oriented spatial structures of a scene. Each GIST descriptor is a
512-dimensional vector. We further use PCA to project GIST features
into 30 dimensions. Finally, we obtain $1,800$ ``documents'', each
of which contains regions as observations. Each region now is represented
by a 30-dimensional vector. We now can perform clustering regions
in every image since they are visually correlated. In the next level
of clustering, we can cluster images into scene categories.
\begin{table}
\caption{Clustering performance for LabelMe
dataset.}
\label{tab:clustering_metrics}
\centering
\begin{tabular}{lcccc}
\toprule
Methods & \textcolor{black}{NMI} & ARI & \textcolor{black}{AMI} & \textcolor{black}{Time (s) } \\
\midrule
\textcolor{black}{K-means} & 0.349 & 0.237 & 0.324 & \textbf{0.3}\\
TSK-means & 0.236 & 0.112 & 0.22 & 218\\
MC2 & 0.315 & 0.206 & 0.273 & 4.2\\
\textbf{\textcolor{black}{MWM}} & 0.373 & 0.263 & 0.352 & 332 \\
\textbf{MWMS} & \textbf{0.391} & \textbf{0.284} & \textbf{0.368} & 544 \\
\bottomrule
\end{tabular}
\end{table}

\comment{
\begin{table}
\begin{centering}
\begin{tabular}{|c|c|c|c|c|}
\hline 
Methods & \textcolor{black}{NMI} & ARI & \textcolor{black}{AMI} & \textcolor{black}{Time (s) }\tabularnewline
\hline 
\hline 
\textcolor{black}{K-means} & 0.349 & 0.237 & 0.324 & \textbf{0.3}\tabularnewline
\hline 
TSK-means & 0.236 & 0.112 & 0.22 & 218\tabularnewline
\hline 
MC2-SVI & 0.315 & 0.206 & 0.273 & 4.2\tabularnewline
\hline 
\textbf{\textcolor{black}{MWM}} & 0.373 & 0.263 & 0.352 & 332 \tabularnewline
\hline 
\textbf{MWMS} & \textbf{0.391} & \textbf{0.284} & \textbf{0.368} & 544 \tabularnewline
\hline 
\end{tabular}
\par\end{centering}

\caption{\label{tab:clustering_metrics}Clustering performance for LabelMe
dataset.}
\end{table}}

\textbf{StudentLife dataset} is a large dataset frequently
used in pervasive and ubiquitous computing research. Data signals consist of
multiple channels  (e.g., WiFi signals, Bluetooth scan, etc.), which are
collected from smartphones of 49 students at Dartmouth
College over a 10-week spring term in 2013. However, in our experiments,
we use only WiFi signal strengths. We applied a similar procedure
described in \cite{nguyen_nguyen_venkatesh_phung_icpr16mcnc} to pre-process the
data. We aggregate the number of scans by each Wifi access point and
select 500 Wifi Ids with the highest frequencies. Eventually, we obtain
49 ``documents'' with totally approximately $4.6$ million 500-dimensional
data points.

\textbf{Experimental results.} To quantitatively evaluate our proposed
methods, we compare our algorithms with several base-line methods:
K-means, three-stage K-means (TSK-means) as described in the Supplement, 
MC2-SVI without context \cite{Viet-2016}. Clustering performance in Table
\ref{tab:clustering_metrics} is evaluated with the image clustering
problem for \emph{LabelMe dataset}. With \emph{K-means}, we
average all data points to obtain a single vector for each images.
K-means needs much less time to run since the number of data points
is now reduced to $1,800$. For MC2-SVI, we used stochastic varitational and 
a parallelized Spark-based implementation in \cite{Viet-2016} to carry out 
experiments. This implementation has the advantage of making use of all of 16 cores on the test machine. 
The running time for MC2-SVI is reported after scanning one epoch.
In terms of clustering accuracy, MWM and MWMS algorithms 
perform the best.

Fig. \ref{fig:labelme_clusters} demonstrates five representative
image clusters with six randomly chosen images in each (on the right)
which are discovered by our MWMS algorithm. We also accumulate labeled
tags from all images in each cluster to produce the tag-cloud on the
left. These tag-clouds can be considered as visual ground truth of
clusters. Our algorithm can group images into clusters which are consistent
with their tag-clouds.

We use StudentLife dataset to demonstrate the capability of multilevel
clustering with large-scale datasets. This dataset not only contains
a large number of data points but presents in high dimension. Our
algorithms need approximately 1 hour to perform multilevel
clustering on this dataset. Fig. \ref{fig:SL-graph} presents two
levels of clusters discovered by our algorithms. The innermost (blue)
and outermost (green) rings depict local and global clusters respectively.
Global clusters represent groups of students while local clusters
shared between students (``documents'') may be used to infer
locations of students' activities. From these clusteing we can dissect
students' shared location (activities), e.g. Student 49 (\emph{U49})
mainly takes part in activity location 4 (\emph{L4}). 

%% file: arxiv_nested_appendix_a.tex
In this appendix, we collect relevant information on the Wasserstein metric and 
Wasserstein barycenter problem, which were introduced in Section \ref{Section:prelim} in the paper. 
For any Borel map $g: \Theta \to \Theta$ and probability measure $G$ on $\Theta$, the 
push-forward measure of $G$ through $g$, denoted by $g\#G$, is defined by the condition
that
$\int \limits_{\Theta}{f(y)}d(g\#G)(y)=\int \limits_{\Theta}{f(g(x))}d G(x)$ for every 
continuous bounded function $f$ on $\Theta$.
\paragraph{Wasserstein metric} \label{Section:Append_Wasserstein_metric}
When $G = \sum \limits_{i=1}^{k}{p_{i}\delta_{\theta_{i}}}$ and $G'=\sum  \limits_{i=1}^{k'}{p_{i}'\delta_{\theta_{i}'}}$ 
are discrete measures with finite support, i.e., $k$ and $k'$ are finite, 
the Wasserstein distance of order $r$ between $G$ and $G'$ can be represented as
\begin{eqnarray}
W_{r}^{r}(G,G') = \min \limits_{T \in \Pi(G,G')} \langle T,M_{G,G'} \rangle \label{eqn:Wasserstein_computation}
\end{eqnarray}
where we have
\begin{eqnarray}
\Pi(G,G') = \left\{T \in \mathbb{R}_{+}^{k \times k'}: T\mathbbm{1}_{k'}=\vec{p}, \ T\mathbbm{1}_{k}=\vec{p}'\right\} \nonumber
\end{eqnarray}
such that $\vec{p}=(p_{1},\ldots,p_{k})^{T}$ and $\vec{p}'=(p_{1}',\ldots,p_{k'}')^{T}$, 
$M_{G,G'} = \left\{\|\theta_{i}-\theta_{j}'\|\right\}_{i,j} \in \mathbb{R}_{+}^{k \times k'}
$ is the cost matrix, i.e. matrix of pairwise distances of elements between $G$ and $G'$, and 
$\langle A, B \rangle =  \text{tr}(A^{T}B)$ is the Frobenius dot-product of matrices. The 
optimal $T \in \Pi(G,G')$ in optimization problem \eqref{eqn:Wasserstein_computation} is called 
the optimal coupling of $G$ and $G'$, representing the \textbf{optimal transport} between these two measures. 
When $k=k'$, 
the complexity of best algorithms for finding the optimal transport is $O(k^{3}\log k)$. 
Currently, \cite{Cuturi-2013} proposed a regularized version of 
\eqref{eqn:Wasserstein_computation} based on Sinkhorn distance where the complexity of 
finding an approximation of the optimal transport is $O(k^{2})$. Due to its favorably
fast computation, throughout the paper we shall utilize Cuturi's algorithm to compute the Wasserstein 
distance between $G$ and $G_{'}$ as well as their optimal transport in 
\eqref{eqn:Wasserstein_computation}.
\paragraph{Wasserstein barycenter} \label{Section:Append_Wasserstein_barycenter}
As introduced in Section \ref{Section:prelim} in the paper, 
for any probability measures $P_{1}, P_{2}, \ldots, P_{N} \in \mathcal{P}_{2}(\Theta)$, their Wasserstein barycenter $\overline{P}_{N,\lambda}$ is such that
\begin{eqnarray}
\overline{P}_{N,\lambda}=\mathop {\arg \min}\limits_{P \in \mathcal{P}_{2}(\Theta)}{\sum \limits_{i=1}^{N}{\lambda_{i}W_{2}^{2}(P,P_{i})}} \nonumber
\end{eqnarray} 
where $\lambda \in \Delta_{N}$ denote weights associated with $P_{1},\ldots,P_{N}$. 
According to \citep{Carlier-2011}, $P_{N,\lambda}$ can be obtained as a solution to 
so-called multi-marginal optimal transporation problem. In fact, if we denote $T_{k}^{1}$ as 
the measure preseving map from $P_{1}$ to $P_{k}$, i.e., 
$P_{k}=T_{k}^{1} \# P_{1}$, for any $1 \leq k \leq N$, then \begin{eqnarray}
\overline{P}_{N,\lambda}=\biggr(\sum \limits_{k=1}^{N}{\lambda_{k}T_{k}^{1}}\biggr)\# P_{1}. \nonumber
\end{eqnarray}
Unfortunately, the forms of the maps $T_{k}^{1}$ are analytically intractable, especially 
if no special constraints on $P_{1}, \ldots, P_{N}$ are imposed.

Recently, \citep{Anderes-2015} studied the Wasserstein barycenters $\overline{P}_{N,
\lambda}$ when $P_{1}, P_{2}, \ldots, P_{N}$ are finite discrete measures and $\lambda=
\biggr(1/N,\ldots,1/N\biggr)$. They demonstrate the following sharp result (cf. Theorem 2 
in \citep{Anderes-2015}) regarding the number of atoms of $\overline{P}_{N,\lambda}$
\begin{customthm}{A.1} \label{theorem:upperbound_barycenter} There exists a Wasserstein 
barycenter $\overline{P}_{N,\lambda}$ such that $\text{supp}(\overline{P}_{N,\lambda}) \leq \sum \limits_{i=1}^{N}{s_{i}}-N+1$.
\end{customthm}
Therefore, when $P_{1},\ldots, P_{N}$ are indeed finite discrete measures and the weights 
are uniform, the problem of finding Wasserstein barycenter $\overline{P}_{N,\lambda}$ over 
the (computationally large) space $\mathcal{P}_{2}(\Theta)$ is 
reduced to a search over a smaller
space $\mathcal{O}_{l}(\Theta)$ where $l=\sum \limits_{i=1}^{N}{s_{i}-N+1}$.

%% file: arxiv_nested_appendix_b.tex
In this appendix, we provide proofs for the remaining results in the paper. 
We start by giving a proof for the transition 
from multilevel Wasserstein means objective function \eqref{eqn:multilevel_Kmeans_typeone} to objective function 
\eqref{eqn:multilevel_K_means_typeone_first} in Section \ref{Section:multilevel_kmeans} in the paper. 
All the notations in this appendix are similar to those in the main text. 
For each closed subset $\mathcal{S} \subset \mathcal{P}_{2}(\Theta)$, 
denote the Voronoi region generated by $\mathcal{S}$ on the space 
$\mathcal{P}_{2}(\Theta)$ by the collection of subsets
$\{ V_P \}_{P \in \mathcal{S}}$, 
where $V_P := \{Q \in \mathcal{P}_{2}(\Theta) : W_{2}^{2}(Q,P) = 
\mathop {\min }\limits_{G \in \mathcal{S}} W_{2}^{2}(Q,G)\}$. We define the projection mapping $\pi _\mathcal{S}$ as: $\pi _{\mathcal{S}} :\mathcal{P}_{2}(\Theta)  \to \mathcal{S}$ 
where $\pi _{\mathcal{S}} (Q) = P$ as $Q \in V_{P}$. Note that, for any $P_{1}, P_{2} \in \mathcal{S}$ such that $V_{P_{1}}$ and $V_{P_{2}}$ share the boundary, the values of $\pi_{S}$ at the elements in that boundary can be chosen to be either $P_{1}$ or $P_{2}$. Now, we start with the following useful lemmas.

\setcounter{lemma}{1}
\begin{customlem}{B.1} \label{lemma:one} For any closed subset $\mathcal{S}$ on $\mathcal{P}_{2}(\Theta)$, if $\mathcal{Q} \in \mathcal{P}_{2}(\mathcal{P}_{2}(\Theta))$, then $E_{X \sim \mathcal{Q}} (d_{W_{2}}^{2}(X,\mathcal{S})) = W_2^2 (\mathcal{Q},\pi _\mathcal{S} \# \mathcal{Q})$ where $d_{W_{2}}^{2}(X,\mathcal{S})=\inf \limits_{P \in \mathcal{S}}{W_{2}^{2}(X,P)}$.
\end{customlem}
\begin{proof}
For any element $\pi \in \Pi (\mathcal{Q},\pi _\mathcal{S} \# \mathcal{Q})$:
\begin{eqnarray}
\int{W_{2}^{2}(P,G)} d\pi (P,G) & \geq &  \int {d_{W_{2}}^{2}(P,\mathcal{S})} d\pi (P,G) \nonumber \\
& = & \int {d_{W_{2}}^{2}(P,\mathcal{S})} d\mathcal{Q}(P) \nonumber \\
& = &  E_{X \sim \mathcal{Q}} (d_{W_{2}}^{2}(X,\mathcal{S})) \nonumber
\end{eqnarray}
where the integrations in the first two terms range over $\mathcal{P}_{2}(\Theta)  \times \mathcal{S}$ while that in the final term ranges over $\mathcal{P}_{2}(\Theta)$. Therefore, we obtain 
\begin{eqnarray}
W_2^2 (\mathcal{Q},\pi _\mathcal{S} \# \mathcal{Q}) & = & \mathop {\inf } \int\limits_{\mathcal{P}_{2}(\Theta)  \times \mathcal{S}} {W_{2}^{2}(P,G)} d\pi (P,G) \nonumber \\
& \geq & E_{X \sim \mathcal{Q}} (d_{W_{2}}^{2}(X,\mathcal{S})) \label{eqn:lemmaequationone}
\end{eqnarray}
where the infimum in the first equality ranges over all $\pi \in \Pi (\mathcal{Q},\pi _\mathcal{S} \# \mathcal{Q})$. 

On the other hand, let ${\displaystyle g:\mathcal{P}_{2}(\Theta)  \to \mathcal{P}_{2}(\Theta) \times \mathcal{S}}$ 
such that $g(P)=(P,\pi_{\mathcal{S}}(P))$ for all $P \in \mathcal{P}_{2}(\Theta)$. Additionally, let 
$\mu _{\pi _\mathcal{S} } = g \# \mathcal{Q}$, the push-forward measure of $\mathcal{Q}$ under mapping $g$. It is clear that $\mu _{\pi _\mathcal{S}}$ is a coupling between $\mathcal{Q}$ and $\pi _\mathcal{S} \# \mathcal{Q}$. 
Under this construction, we obtain for any $X \sim \mathcal{Q}$ that
\begin{eqnarray}
 E\left(W_{2}^{2}(X,\pi _\mathcal{S} (X))\right) & = & \int {W_{2}^{2}(P,G) } d\mu _{\pi _\mathcal{S}} (P,G) \nonumber \\
 & \geq & \mathop {\inf } \int {W_{2}^{2}(P,G)} d\pi (P,G) \nonumber \\
& = & W_2^2 (\mathcal{Q},\pi _\mathcal{S} \# \mathcal{Q}) \label{eqn:lemmaequationtwo}
\end{eqnarray}
where the infimum in the second inequality ranges over all $\pi  \in \Pi (\mathcal{Q},\pi _\mathcal{S} \# \mathcal{Q})$ and the integrations range over $\mathcal{P}_{2}(\Theta) \times \mathcal{S}$. Now, from the definition of $\pi_{\mathcal{S}}$
\begin{eqnarray}
E(W_{2}^{2}(X,\pi _\mathcal{S} (X))) & = & \int {W_{2}^{2}(P,\pi _\mathcal{S}(P))}d\mathcal{Q}(P) \nonumber \\
& = & \int {d_{W_{2}}^{2}(P,\mathcal{S})} d\mathcal{Q}(P) \nonumber \\
& = & E(d_{W_{2}}^{2}(X,\mathcal{S})) \label{eqn:lemmaequationthree}
\end{eqnarray}
where the integrations in the above equations range over $\mathcal{P}_{2}(\Theta)$. By combining \eqref{eqn:lemmaequationtwo} and \eqref{eqn:lemmaequationthree}, we would obtain that
\begin{eqnarray}
E_{X \sim \mathcal{Q}} (d^{2}_{W_{2}}(X,\mathcal{S})) \ge W_2^2 (\mathcal{Q},\pi _\mathcal{S} \# \mathcal{Q}). \label{eqn:lemmaequationfourth}
\end{eqnarray}
From \eqref{eqn:lemmaequationone} and \eqref{eqn:lemmaequationfourth}, it is straightforward that $E_{X \sim Q} (d(X,S)^2 ) = W_2^2 (Q,\pi _S  \# Q)$. Therefore, we achieve the conclusion of the lemma.
\end{proof}

\begin{customlem}{B.2} \label{lemma:two}
For any closed subset $\mathcal{S} \subset \mathcal{P}_{2}(\Theta)$ and $\mu \in \mathcal{P}_{2}(\mathcal{P}_{2}(\Theta))$ 
with $\text{supp}(\mu) \subseteq \mathcal{S}$, 
there holds 
$W_2^2 (\mathcal{Q},\mu ) \ge W_2^2 (\mathcal{Q},\pi _\mathcal{S} \# \mathcal{Q})$ for any $\mathcal{Q} \in \mathcal{P}_{2}(\mathcal{P}_{2}(\Theta))$.
\end{customlem}
\begin{proof}
Since $\text{supp}(\mu) \subseteq \mathcal{S}$, it is clear that $W_2^2 (\mathcal{Q},\mu) = {\displaystyle \mathop {\inf }\limits_{\pi  \in \Pi (\mathcal{Q},\mu )} \int\limits_{\mathcal{P}_{2}(\Theta) \times \mathcal{S}} {W_{2}^{2}(P,G)} d\pi (P,G)}$.\\
Additionally, we have
\begin{eqnarray}
\int {W_{2}^{2}(P,G)} d\pi (P,G) & \geq & \int {d_{W_{2}}^{2}(P,\mathcal{S})} d\pi (P,G) \nonumber \\
& = &  \int {d_{W_{2}}^{2}(P,\mathcal{S})} d\mathcal{Q}(P) \nonumber \\
& = & E_{X \sim Q} (d_{W_{2}}^{2}(X,S)) \nonumber \\
& = & W_2^2 (\mathcal{Q},\pi _\mathcal{S} \# \mathcal{Q}) \nonumber
\end{eqnarray}
where the last inequality is due to Lemma \ref{lemma:one} and the integrations in the first two terms range over $\mathcal{P}_{2}(\Theta) \times \mathcal{S}$ while that in the final term ranges over $\mathcal{P}_{2}(\Theta)$. Therefore, we achieve the conclusion of the lemma.
\end{proof}
Equipped with Lemma \ref{lemma:one} and Lemma \ref{lemma:two}, 
we are ready to establish 
the equivalence between multilevel Wasserstein means objective function (5) and objective function (4) in Section \ref{Section:multilevel_kmeans} in the main text.
\begin{customlem}{B.3} \label{proposition:Wassersteinequivalence}
For any given positive integers $m$ and $M$, we have
\begin{eqnarray}
A : = \inf \limits_{\Hcal \in \mathcal{E}_{M}(\mathcal{P}_{2}(\Theta))} W_{2}^{2}(\Hcal,\dfrac{1}{m}\mathop {\sum }\limits_{j=1}^{m}{\delta_{G_{j}}}) \nonumber \\
= \dfrac{1}{m}\inf \limits_{\Hbold = (H_{1},\ldots,H_{M})}\sum \limits_{j=1}^{m} d_{W_{2}}^{2}(G_{j},\Hbold) := B. \nonumber
\end{eqnarray}
\end{customlem}
\begin{proof}
Write $\mathcal{Q}=\dfrac{1}{m}\mathop {\sum }\limits_{j=1}^{m}{\delta_{G_{j}}}$. From the definition of $B$, for any $\epsilon>0$, we can find $\overline{\Hbold}$ such that 
\begin{eqnarray}
B & \geq & \dfrac{1}{m}\sum \limits_{j=1}^{m} d_{W_{2}}^{2}(G_{j},\overline{\Hbold}) - \epsilon \nonumber \\
& = & E_{X \sim \mathcal{Q}}(d_{W_{2}}^{2}(X,\overline{\Hbold})) - \epsilon \nonumber \\
& = & W_{2}^{2}(\mathcal{Q},\pi_{\overline{\Hbold}} \# \mathcal{Q}) {\bf - \epsilon} \nonumber \\
& \geq & A -\epsilon \nonumber
\end{eqnarray}
where the second equality in the above display
is due to Lemma \ref{lemma:one} while the last 
inequality is from the fact that $\pi_{\overline{\Hbold}} \# \mathcal{Q}$ is a discrete probability measure in $\mathcal{P}_{2}(\mathcal{P}_{2}(\Theta))$ with exactly $M$ support points. Since the inequality in the above display holds for any $\epsilon$, it implies that $B \geq A$. On the other hand, from the formation of $A$, for any $\epsilon>0$, we also can find $\Hcal' \in \mathcal{E}_{M}(\mathcal{P}_{2}(\Theta))$ such that
\begin{eqnarray}
A & \geq & W_{2}^{2}(\Hcal',\mathcal{Q}) - \epsilon \nonumber \\
& \geq & W_{2}^{2}(\mathcal{Q},\pi_{\Hbold'} \# \mathcal{Q}) - \epsilon \nonumber \\
& = & \dfrac{1}{m}\sum \limits_{j=1}^{m} d_{W_{2}}^{2}(G_{j},\Hbold') -\epsilon \nonumber \\
& \geq & B - \epsilon \nonumber
\end{eqnarray}
where  $\Hbold' = \text{supp}(\Hcal')$, the second inequality is due to Lemma \ref{lemma:two}, and the third equality is due to Lemma \ref{lemma:one}. Therefore, it means that $A \geq B$. We achieve the conclusion of the lemma. 
\end{proof}
\begin{customprop}{B.4} \label{lemma:equivalence_multilevels_Kmeans}
For any positive integer numbers $m, M$ and $k_{j}$ as $1 \leq j \leq m$, we denote 
\begin{eqnarray}
C & : = & \mathop {\inf }\limits_{\substack {G_{j} \in \mathcal{O}_{k_{j}}(\Theta) \ \forall 1 \leq j \leq m, \\ \Hcal \in \mathcal{E}_{M}(\mathcal{P}_{2}(\Theta))}}{\mathop {\sum }\limits_{i=1}^{m}{W_{2}^{2}(G_{j},P_{n_{j}}^{j})}} \nonumber \\
& + & W_{2}^{2}(\Hcal,\dfrac{1}{m}\mathop {\sum }\limits_{i=1}^{m}{\delta_{G_{i}}}) \nonumber \\
D & : = & \mathop {\inf }\limits_{\substack {G_{j} \in \mathcal{O}_{k_{j}}(\Theta) \ \forall 1 \leq j \leq m, \\ \Hbold = (H_{1},\ldots,H_{M})}}{\mathop {\sum }\limits_{j=1}^{m}{W_{2}^{2}(G_{j},P_{n_{j}}^{j})}} \nonumber \\
& + & \dfrac{d_{W_{2}}^{2}(G_{j},\Hbold)}{m}. \nonumber
\end{eqnarray}
Then, we have $C = D$.
\end{customprop}
\begin{proof} The proof of this proposition is a straightforward application of Lemma \ref{proposition:Wassersteinequivalence}. Indeed, for each fixed $(G_{1},\ldots,G_{m})$ the infimum w.r.t to $\mathcal{H}$ in $C$ leads to the same infimum w.r.t to $\Hbold$ in $D$, according to Lemma \ref{proposition:Wassersteinequivalence}. Now, by taking the infimum w.r.t to $(G_{1},\ldots,G_{m})$ on both sides, we achieve the conclusion of the proposition.
\end{proof}

In the remainder of the Supplement, we present the proofs for all remaining
theorems stated in the main text.
\paragraph{PROOF OF THEOREM \ref{theorem:local_convergence_multilevel_Kmeans}} 
The proof of this theorem is straightforward from the formulation of Algorithm \ref{alg:multilevels_Wasserstein_means}. 
In fact, for any $G_{j} \in \mathcal{E}_{k_{j}}(\Theta)$ and $\Hbold =(H_{1},\ldots,H_{M})$, we denote the function 
\begin{eqnarray}
f(\vec{G}, \Hbold)=\mathop {\sum }\limits_{j=1}^{m}{W_{2}^{2}(G_{j},P_{n}^{j})}+\dfrac{d_{W_{2}}^{2}(G_{j},\Hbold)}{m} \nonumber
\end{eqnarray}
where $\vec{G}=(G_{1},\ldots,G_{m})$. To obtain the conclusion of this theorem, it is sufficient to demonstrate for any $t \geq 0$ that
\begin{eqnarray}
f(\vec{G}^{(t+1)},\Hbold^{(t+1)}) \leq f(\vec{G}^{(t)},\Hbold^{(t)}). \nonumber
\end{eqnarray}
This inequality comes directly from $f(\vec{G}^{(t+1)},\Hbold^{(t)}) \leq f(\vec{G}^{(t)},\Hbold^{(t)})$, which is due to the Wasserstein barycenter problems to obtain $G_{j}^{(t+1)}$ for $1 \leq j \leq m$, and $f(\vec{G}^{(t+1)},\Hbold^{(t+1)}) \leq f(\vec{G}^{(t+1)},\Hbold^{(t)})$, which is due to the optimization steps to achieve elements $H_{u}^{(t+1)}$ of $\Hbold^{(t+1)}$ as $1 \leq u \leq M$. As a consequence, we achieve the conclusion of the theorem.
\paragraph{PROOF OF THEOREM \ref{theorem:objective_consistency_multilevel_Wasserstein_means}} 
To simplify notation, write
\begin{eqnarray}
L_{\vec{n}}=\mathop {\inf }\limits_{\substack {G_{j} \in \mathcal{O}_{k_{j}}(\Theta), \\ \Hcal \in \mathcal{E}_{M}(\mathcal{P}_{2}(\Theta))}}f_{\vec{n}}(\vec{G},\Hcal), \nonumber \\
L_{0}=\mathop {\inf }\limits_{\substack {G_{j} \in \mathcal{O}_{k_{j}}(\Theta), \\ \Hcal \in \mathcal{E}_{M}(\mathcal{P}_{2}(\Theta))}}f(\vec{G},\Hcal). \nonumber
\end{eqnarray}
For any $\epsilon>0$, from the definition of $L_{0}$, we can find $G_{j} \in \mathcal{O}_{k_{j}}(\Theta)$ and $\Hcal \in \mathcal{E}_{M}(\mathcal{P}(\Theta))$ such that
\begin{eqnarray}
f(\vec{G},\Hcal)^{1/2} \leq L_{0}^{1/2} + \epsilon. \nonumber
\end{eqnarray}
Therefore, we would have
\begin{eqnarray}
L_{\vec{n}}^{1/2}-L_{0}^{1/2} & \leq & L_{n}^{1/2}-f(\vec{G},\Hcal)^{1/2}+\epsilon \nonumber \\
& \leq & f_{\vec{n}}(\vec{G},\Hcal)^{1/2} - f(\vec{G},\Hcal)^{1/2}+\epsilon \nonumber \\
& = & \dfrac{f_{\vec{n}}(\vec{G},\Hcal)-f(\vec{G},\Hcal)}{f_{\vec{n}}(\vec{G},\Hcal)^{1/2}+f(\vec{G},\Hcal)^{1/2}} + \epsilon \nonumber \\
& \leq & \sum \limits_{j=1}^{m}\dfrac{|W_{2}^{2}(G_{j},P_{n_{j}}^{j})-W_{2}^{2}(G_{j},P^{j})|}{W_{2}(G_{j},P_{n_{j}}^{j})+W_{2}(G_{j},P^{j})}+\epsilon \nonumber \\
& \leq & \sum \limits_{j=1}^{m}{W_{2}(P_{n_{j}}^{j},P^{j})}+\epsilon. \nonumber
\end{eqnarray} 
By reversing the direction, we also obtain the inequality $L_{n}^{1/2}-L_{0}^{1/2} \geq \sum \limits_{j=1}^{m}{W_{2}(P_{n_{j}}^{j},P^{j})}-\epsilon$. Hence, $|L_{n}^{1/2}-L_{0}^{1/2}-\sum \limits_{j=1}^{m}{W_{2}(P_{n_{j}}^{j},P^{j})}| \leq \epsilon$ for any $\epsilon>0$. Since $P^{j} \in \mathcal{P}_{2}(\Theta)$ for all $1 \leq j \leq m$, we obtain that $W_{2}(P_{n_{j}}^{j},P^{j}) \to 0$ almost surely as $n_{j} \to \infty$ (see for example Theorem 6.9 in \citep{Villani-2009}). As a consequence, we obtain the conclusion of the theorem.
\paragraph{PROOF OF THEOREM \ref{theorem:convergence_measures_multilevel_Wasserstein_means}} 
For any $\epsilon>0$, we denote
\begin{eqnarray}
\mathcal{A}(\epsilon)=\biggr\{G_{i} \in \mathcal{O}_{k_{i}}(\Theta), \Hcal \in \mathcal{E}_{M}(\mathcal{P}(\Theta)): \nonumber \\
 d(\vec{G},\Hcal,\mathcal{F}) \geq \epsilon\biggr\}. \nonumber
\end{eqnarray}
Since $\Theta$ is a compact set, we also have $\mathcal{O}_{k_{j}}(\Theta)$ and $\mathcal{E}_{M}(\mathcal{P}_{2}(\Theta))$ are compact for any $1 \leq i \leq m$. As a consequence, $\mathcal{A}(\epsilon)$ is also a compact set. For any $(\vec{G},\Hcal) \in \mathcal{A}(\epsilon)$, by the definition of $\mathcal{F}$ we would have $f(\vec{G},\Hcal) > f(\vec{G}^{0},\Hcal^{0})$ for any $(\vec{G}^{0},\Hcal^{0}) \in \mathcal{F}$. Since $\mathcal{A}(\epsilon)$ is compact, it leads to
\begin{eqnarray}
\inf \limits_{(\vec{G},\Hcal) \in A(\epsilon)}{f(\vec{G},\Hcal)} > f(\vec{G}^{0},\Hcal^{0}). \nonumber
\end{eqnarray}
for any $(\vec{G}^{0},\Hcal^{0}) \in \mathcal{F}$. From the formulation of $f_{\vec{n}}$ as in the proof of Theorem \ref{theorem:objective_consistency_multilevel_Wasserstein_means}, 
we can verify that $\lim \limits_{\vec{n} \to \infty} f_{\vec{n}}(\widehat{\vec{G}}^{\vec{n}},\widehat{\Hcal}^{\vec{n}}) = \lim \limits_{\vec{n} \to \infty} f(\widehat{\vec{G}}^{\vec{n}},\widehat{\Hcal}^{\vec{n}})$ almost surely as 
$\vec{n} \to \infty$. Combining this result with that of Theorem \ref{theorem:objective_consistency_multilevel_Wasserstein_means}, 
we obtain $f(\widehat{\vec{G}}^{\vec{n}},\widehat{\Hcal}^{\vec{n}}) \to f(\vec{G}^{0},\Hcal^{0})$ as $\vec{n} \to \infty$ for any $(\vec{G}^{0},\Hcal^{0}) \in \mathcal{F}$. Therefore, for any $\epsilon>0$, as $\vec{n}$ is large enough, we have $d(\vec{\widehat{G}}^{\vec{n}},\widehat{\Hcal}^{\vec{n}},\mathcal{F}) < \epsilon$. As a consequence, we achieve the conclusion regarding the consistency of the mixing measures.

%% file: arxiv_nested_appendix_c.tex
In this appendix, we provide details on the algorithm for the Multilevel Wasserstein means with sharing (MWMS) 
formulation (Algorithm \ref{alg:local_constraint_multilevels_Wasserstein_means}). Recall the MWMS objective function as follows
\begin{eqnarray}
\mathop {\inf }\limits_{\mathcal{S}_{K},G_{j} , \Hbold \in \mathcal{B}_{M,\mathcal{S}_{K}}}{\mathop {\sum }\limits_{j=1}^{m}{W_{2}^{2}(G_{j},P_{n_{j}}^{j})}+\dfrac{d_{W_{2}}^{2}(G_{j},\Hbold)}{m}} \nonumber
\end{eqnarray}
where  $\mathcal{B}_{M,\mathcal{S}_{K}}=\biggr\{G_{j} \in \mathcal{O}_{K}(\Theta), \ \Hbold=(H_{1},\ldots,H_{M}): 
\text{supp}(G_{j})\subseteq \mathcal{S}_{K}\ \forall 1 \leq j \leq m \biggr\}$. 

\setcounter{algorithm}{1}
\begin{algorithm}[tbp]
   \caption{Multilevel Wasserstein Means with Sharing (MWMS)}
   \label{alg:local_constraint_multilevels_Wasserstein_means}
\begin{algorithmic}
   \STATE {\bfseries Input:} Data $X_{j,i}$, $K$, $M$.
   \STATE {\bfseries Output:} global set $S_{K}$, local measures $G_{j}$, and elements $H_{i}$ of $\Hbold$.
   \STATE Initialize $S_{K}^{(0)}=\left\{a_{1}^{(0)},\ldots,a_{K}^{(0)}\right\}$, elements $H_{i}^{(0)}$ of $\Hbold^{(0)}$, and $t = 0$.
   \WHILE{$S_{K}^{(t)},G_{j}^{(t)},H_{i}^{(t)}$ have not converged}
   \STATE 1. Update global set $S_{K}^{(t)}$:
   \FOR{$j=1$ {\bfseries to} $m$}
   \STATE $i_{j} \leftarrow \mathop {\arg \min}\limits_{1 \leq u \leq M}{W_{2}^{2}(G_{j}^{(t)},H_{u}^{(t)})}$.
   \STATE $T^{j} \leftarrow$ optimal coupling of $G_{j}^{(t)}$, $P_{n}^{j}$ (cf. Appendix A).
   \STATE $U^{j} \leftarrow$ optimal coupling of $G_{j}^{(t)}$, $H_{i_{j}}^{(t)}$.
   \ENDFOR
   \FOR{$i=1$ {\bfseries to} $M$}
   \STATE $h_{i}^{(t)} \leftarrow$ atoms of $H_{i}^{(t)}$ with $h_{i,v}^{(t)}$ as v-th column.
   \ENDFOR
   \FOR {$i=1$ {\bfseries to} $K$}
   \STATE $m D \leftarrow m \sum \limits_{u=1}^{m}{\sum \limits_{v=1}^{n_{i}}{T_{i,v}^{u}}}+\sum \limits_{u=1}^{m}{\sum \limits_{v \neq i}{U_{i,v}^{u}}}$.
   \STATE $a_{i}^{(t+1)} \leftarrow \biggr(m \sum \limits_{u=1}^{m}{\sum \limits_{v=1}^{n_{i}}{T_{i,v}^{u}X_{u,v}}}+$\\
$\sum \limits_{u=1}^{m}{\sum \limits_{v}{U_{i,v}^{u}h_{j_{u},v}^{(t)}}}\biggr)/mD$.
	\ENDFOR
	\STATE 2. Update $G_{j}^{(t)}$ for $1 \leq j \leq m$:
	\FOR{$j=1$ {\bfseries to} $m$}
	\STATE $G_{j}^{(t+1)} \leftarrow \mathop {\arg \min}\limits_{G_{j}: \text{supp}(G_{j}) \equiv \mathcal{S}_{K}^{(t+1)}}{W_{2}^{2}(G_{j},P_{n_{j}}^{j})}$ \\
	$+W_{2}^{2}(G_{j},H_{i_{j}}^{(t)})/m$.
	\ENDFOR
   \STATE 3. Update $H_{i}^{(t)}$ for $1 \leq i \leq M$ as Algorithm \ref{alg:multilevels_Wasserstein_means}.
   \STATE 4. $t \leftarrow t+1$.
   \ENDWHILE
\end{algorithmic}
\end{algorithm}

We make the following remarks regarding the initializations and updates of Algorithm \ref{alg:local_constraint_multilevels_Wasserstein_means}: 
\begin{itemize}
\item[(i)] An efficient way to 
initialize global set $S_{K}^{(0)}=\biggr\{a_{1}^{(0)},\ldots,a_{K}^{(0)}\biggr\} \in 
\mathbb{R}^{d \times K}$ is to perform $K$-means on the whole data set $X_{j,i}$ for $1 
\leq j \leq m, 1 \leq i \leq n_{j}$; 
\item[(ii)] The updates $a_{j}^{(t+1)}$ are indeed the solutions 
of the following optimization problems
\begin{eqnarray}
\inf \limits_{a_{j}^{(t)}}\biggr\{\sum \limits_{l=1}^{m}{W_{2}^{2}(G_{l}^{(t)},P_{n}^{l})}+\dfrac{\sum \limits_{l=1}^{m}{W_{2}^{2}(G_{l}^{(t)},H_{i_{l}}^{(t)})}}{m}\biggr\}, \nonumber
\end{eqnarray}
which is equivalent to find $a_{j}^{(t)}$ to optimize 
\begin{eqnarray}
m \sum \limits_{u=1}^{m}{\sum \limits_{v=1}^{n_{j}}{T_{j,v}^{u}\|a_{j}^{(t)}-X_{u,v}\|^{2}}} \nonumber \\
+\sum \limits_{u=1}^{m}{\sum \limits_{v}{U_{j,v}^{u}\|a_{j}^{(t)}-h_{i_{j},v}^{(t)}||^{2}}}.\nonumber
\end{eqnarray}
where $T^{j}$ is an optimal coupling of $G_{j}^{(t)}$, $P_{n}^{j}$ and $U^{j}$ is an optimal coupling of $G_{j}^{(t)}$, $H_{i_{j}}^{(t)}$. By taking the first order derivative of the above function with respect to $a_{j}^{(t)}$, we quickly achieve $a_{j}^{(t+1)}$ as the closed form minimum of that function;
\item[(iii)] Updating the local weights of $G_{j}^{(t+1)}$ is equivalent to updating $G_{j}^{(t+1)}$ as 
the atoms of $G_{j}^{(t+1)}$ are known to stem from $S_{K}^{(t+1)}$. 
\end{itemize}
Now, similar to 
Theorem 3.1 in the main text, we also have the following theoretical 
guarantee regarding the behavior of Algorithm \ref{alg:local_constraint_multilevels_Wasserstein_means} as follows
\begin{customthm}{C.1} \label{theorem:local_convergence_local_constraint_Kmeans} Algorithm 
\ref{alg:local_constraint_multilevels_Wasserstein_means} monotonically decreases the 
objective function of  
the MWMS formulation.
\end{customthm}
\begin{proof}
The proof is quite similar to the proof of Theorem \ref{theorem:local_convergence_multilevel_Kmeans}. In fact, recall from the proof of Theorem \ref{theorem:local_convergence_multilevel_Kmeans} that for any $G_{j} \in \mathcal{E}_{k_{j}}(\Theta)$ and $\Hbold =(H_{1},\ldots,H_{M})$ we denote the function 
\begin{eqnarray}
f(\vec{G}, \Hbold)=\mathop {\sum }\limits_{j=1}^{m}{W_{2}^{2}(G_{j},P_{n}^{j})}+\dfrac{d_{W_{2}}^{2}(G_{j},\Hbold)}{m} \nonumber
\end{eqnarray}
where $\vec{G}=(G_{1},\ldots,G_{m})$. Now it is sufficient to demonstrate for any $t \geq 0$ that
\begin{eqnarray}
f(\vec{G}^{(t+1)},\Hbold^{(t+1)}) \leq f(\vec{G}^{(t)},\Hbold^{(t)}). \nonumber
\end{eqnarray}
where the formulation of $f$ is similar as in the proof of Theorem \ref{theorem:local_convergence_multilevel_Kmeans}. 
Indeed, by the definition of Wasserstein distances, we have
\begin{eqnarray}
E = m f(\vec{G}^{(t)},\Hbold^{(t)})  = \nonumber \\
 \sum \limits_{u=1}^{m}{\sum \limits_{j,v}{mT_{j,v}^{u}\|a_{j}^{(t)}-X_{u,v}\|^{2}} +  U_{j,v}^{u}\|a_{j}^{(t)}-h_{i_{u},v}^{(t)}\|^{2}}. \nonumber
\end{eqnarray}
Therefore, the update of $a_{i}^{(t+1)}$ from Algorithm \ref{alg:local_constraint_multilevels_Wasserstein_means} 
leads to
\begin{eqnarray}
E & \geq &  \sum \limits_{u=1}^{m}{\sum \limits_{j,v}{mT_{j,v}^{u}\|a_{j}^{(t+1)}-X_{u,v}\|^{2}}} \nonumber \\
& + & U_{j,v}^{u}\|a_{j}^{(t+1)}-h_{i_{u},v}^{(t)}\|^{2} \nonumber \\
& \geq & m \sum \limits_{j=1}^{m}{W_{2}^{2}(G_{j}^{(t)'},P_{n}^{j})}+\sum \limits_{j=1}^{m}{W_{2}^{2}(G_{j}^{(t)'},H_{i_{j}}^{(t)})} \nonumber \\
& \geq & m \sum \limits_{j=1}^{m}{W_{2}^{2}(G_{j}^{(t)'},P_{n}^{j})}+\sum \limits_{j=1}^{m}{d_{W_{2}}^{2}(G_{j}^{(t)'},\Hbold^{(t)})} \nonumber \\
& = & mf(\vec{G'}^{(t)},\Hbold^{(t)}) \nonumber
\end{eqnarray}
where $\vec{G'}^{(t)}=(G_{1}^{(t)'},\ldots,G_{m}^{(t)'})$, $G_{j}^{(t)'}$ are formed by replacing the atoms of $G_{j}^{(t)}$ by the 
elements of $S_{K}^{(t+1)}$, noting that $\text{supp}(G_{j}^{(t)'}) \subseteq  \mathcal{S}_{K}^{(t+1)}$ as $1 \leq j \leq m$, and 
the second inequality comes directly from the definition of Wasserstein distance. Hence, we obtain
\begin{eqnarray}
f(\vec{G}^{(t)},\Hbold^{(t)})\geq f(\vec{G'}^{(t)},\Hbold^{(t)}). \label{eqn:theorem_constraint_Wasserstein_means_first}
\end{eqnarray}
From the formation of $G_{j}^{(t+1)}$ as $1 \leq j \leq m$, we get
\begin{eqnarray}
\sum \limits_{j=1}^{m}{d_{W_{2}}^{2}(G_{j}^{(t+1)},\Hbold^{(t)})} \leq \sum \limits_{j=1}^{m}{d_{W_{2}}^{2}(G_{j}^{(t)'},\Hbold^{(t)})}. \nonumber
\end{eqnarray}
Thus, it leads to 
\begin{eqnarray}
f(\vec{G'}^{(t)},\Hbold^{(t)}) \geq f(\vec{G}^{(t+1)},\Hbold^{(t)}). \label{eqn:theorem_constraint_Wasserstein_means_second}
\end{eqnarray}
Finally, from the definition of $H_{1}^{(t+1)},\ldots,H_{M}^{(t+1)}$, we have
\begin{eqnarray}
f(\vec{G}^{(t+1)},\Hbold^{(t)}) \geq f(\vec{G}^{(t+1)},\Hbold^{(t+1)}). \label{eqn:theorem_constraint_Wasserstein_means_third}
\end{eqnarray}
By combining \eqref{eqn:theorem_constraint_Wasserstein_means_first}, \eqref{eqn:theorem_constraint_Wasserstein_means_second}, and \eqref{eqn:theorem_constraint_Wasserstein_means_third},  we arrive at the conclusion of the theorem.
\end{proof}

%% file: arxiv_nested_appendix_d.tex
In this appendix, we offer details on the data generation processes 
utilized in the simulation studies presented in Section \ref{Section:data_analysis} in the main text. 
The notions of $m, n, d, M$ are given in the main text. 
Let $K_i$ be the number of supporting atoms of $H_i$ and $k_{j}$
the number of atoms of $G_{j}$. For any $d \geq 1$, we denote $\vec{1}_{d}$ to be d dimensional vector with all components to be 1. Furthermore, $\mathcal{I}_{d}$ is an identity matrix with d dimensions. 

\paragraph{Comparison metric (Wasserstein distance to truth)}
\begin{eqnarray}
\text{W}:= \frac{1}{m}\sum_{j=1}^m W_2(\hat G_j, G_j) + d_{\mathcal{M}}(\hat {\Hbold}, \Hbold) \nonumber
\end{eqnarray}
where $\hat{\Hbold} := \{\hat H_1,\ldots,\hat H_M\}$, $\Hbold := \{H_1,\ldots,H_M\}$ and $d_{\mathcal{M}}(\hat H, H)$ is a minimum-matching distance \cite{tang2014understanding, Nguyen-2015}:
$$d_{\mathcal{M}}(\hat{\Hbold}, \Hbold) := \max\{\overline{d}(\hat{\Hbold}, \Hbold), \overline{d}(\Hbold, \hat{\Hbold})\}$$
where
$$\overline{d}(\hat{\Hbold}, \Hbold) := \max\limits_{1 \leq i \leq M}\,\,\min\limits_{1 \leq j \leq M} \,\, W_2(H_{i},\hat H_{j}).$$

\paragraph{Multilevel Wasserstein means setting}
The global clusters are generated as follows:
\begin{eqnarray*}
\left.\begin{aligned}
& \text{means for atoms } \mu_i := 5(i-1), i=1,\ldots, M. \\
& \text{atoms of } H_i: \phi_{ij} \thicksim \mathcal{N}(\mu_i \vec{1}_d, \mathcal{I}_d), j=1,\ldots, K_i.\\
& \text{weights of atoms: } \pi_i \thicksim \text{Dir}(\vec{1}_{K_i}). \\
& \text{Let } H_i := \sum_{j=1}^{K_i} \pi_{ij}\delta_{\phi_{ij}}.
\end{aligned}\right.
\end{eqnarray*}
For each group $j=1,\ldots,m$, generate local measures and data as follows:
\begin{eqnarray*}
\left.\begin{aligned}
& \text{pick cluster label } z_j \thicksim \text{Unif}( \{1,\ldots, M\}).\\
& \text{mean for atoms}: \tau_{ji} \thicksim H_{z_j}, i=1,\ldots, k_j.\\
& \text{atoms of } G_j: \theta_{ji} \thicksim \mathcal{N}(\tau_{ji},\mathcal{I}_d), i=1,\ldots, k_j.\\
& \text{weights of atoms } p_j \thicksim \text{Dir}(\vec{1}_{k_j}). \\
& \text{Let } G_j := \sum_{i=1}^{k_j} p_{ji}\delta_{\theta_{ji}}. \\
& \text{data mean } \mu_i \thicksim G_j, i=1,\ldots, n_j. \\
& \text{observation } X_{j,i} \thicksim \mathcal{N}(\mu_i,\mathcal{I}_d).
\end{aligned}\right.
\end{eqnarray*}
For the case of non-constrained variances, 
the variance to generate atoms $\theta_{ji}$ of $G_{j}$ is set to be proportional to global cluster label $z_{j}$ assigned to $G_{j}$.

\textbf{Multilevel Wasserstein means with sharing setting}\\
The global clusters are generated as follows:
\begin{eqnarray*}
\left.\begin{aligned}
& \text{means for atoms } \mu_i := 5(i-1), i=1,\ldots, M.\\
& \text{atoms of } H_i: \phi_{ij} \thicksim \mathcal{N}(\mu_i \vec{1}_d, \mathcal{I}_d), j=1,\ldots, K_i.\\
& \text{weights of atoms } \pi_i \thicksim \text{Dir}(\vec{1}_{K_i}).\\
& \text{Let }
H_i := \sum_{j=1}^{K_i} \pi_{ij}\delta_{\phi_{ij}}. \\
\end{aligned}\right.
\end{eqnarray*}
For each shared atom $k=1,\ldots,K$:
\begin{eqnarray*}
\left.\begin{aligned}
& \text{pick cluster label } z_k \thicksim \text{Unif}( \{1,\ldots ,M \}).\\
& \text{mean for atoms}: \tau_k \thicksim H_{z_k}.\\
& \text{atoms of } S_K: \theta_k \thicksim \mathcal{N}(\tau_k,\mathcal{I}_d).
\end{aligned}\right.
\end{eqnarray*}
For each group $j=1,\ldots,m$ generate local measures and data as follows:
\begin{eqnarray*}
\left.\begin{aligned}
& \text{pick cluster label } \tilde z_j \thicksim \text{Unif}( \{1,\ldots ,M \}).\\
& \text{select shared atoms } s_j = \{k:z_k=\tilde z_j\}.\\
& \text{weights of atoms } p_{s_j} \thicksim \text{Dir}(\vec{1}_{|s_j|});
\quad
G_j := \sum_{i \in s_j} p_{i}\delta_{\theta_{i}}.\\
& \text{data mean } \mu_i \thicksim G_j, i=1,\ldots,n_j.\\
& \text{observation } X_{j,i} \thicksim \mathcal{N}(\mu_i,\mathcal{I}_d).
\end{aligned}\right.
\end{eqnarray*}
For the case of non-constrained variances, the variance to generate atoms $\theta_{i}$ of $G_{j}$ where $i \in s_{j}$ is set to be proportional to global cluster label $\tilde z_{j}$ assigned to $G_{j}$.


\textbf{Three-stage K-means}
First, we estimate $G_j$ for each group $1 \leq j \leq m$ by using K-means 
algorithm with $k_{j}$ clusters.
Then, we cluster labels using K-means algorithm with $M$ clusters based on the collection of  all atoms of $G_j$s.
Finally, we estimate the atoms of each $H_i$ via K-means algorithm with exactly $L$ clusters for each 
group of local atoms. Here, $L$ is some given threshold being used in Algorithm \ref{alg:multilevels_Wasserstein_means} in 
Section \ref{Section:multilevel_kmeans} in the main text to speed up the computation (see final remark regarding Algorithm \ref{alg:multilevels_Wasserstein_means} in Section \ref{Section:multilevel_kmeans}). 
The three-stage K-means algorithm is summarized in Algorithm \ref{alg:three_stages_K_means}.
\setcounter{algorithm}{2}
\begin{algorithm}
   \caption{Three-stage K-means}
   \label{alg:three_stages_K_means}
\begin{algorithmic}
   \STATE {\bfseries Input:} Data $X_{j,i}$, $k_{j}$, $M$, $L$.
   \STATE {\bfseries Output:} local measures $G_{j}$ and global elements $H_{i}$ of $\Hbold$.
   \STATE {\emph{Stage 1}}
   \FOR{$j=1$ {\bfseries to} $m$}
   \STATE $G_{j} \leftarrow$ $k_{j}$ clusters of group j with K-means (atoms as centroids and weights as label frequencies).
   \ENDFOR
   	\STATE $\mathcal{C} \leftarrow$ collection of all atoms of $G_{j}$.
   \STATE {\emph{Stage 2}}
   	\STATE $\left\{D_{1},\ldots,D_{M}\right\} \leftarrow$ $M$ clusters from K-means on $\mathcal{C}$.
   	\STATE {\emph{Stage 3}}
   \FOR{$i=1$ {\bfseries to} $M$}
   \STATE $H_i \leftarrow$ $L$ clusters of $D_i$ with K-means (atoms as centroids and weights as label frequencies).
   \ENDFOR 
\end{algorithmic}
\end{algorithm}